\pdfoutput=1
\documentclass[11pt]{article}
\usepackage{amsfonts}
\usepackage{amsmath}
\usepackage{booktabs}
\usepackage{makecell} 
\usepackage{enumitem}

\setlist[itemize]{
  topsep=2pt,
  itemsep=2pt,
  parsep=0pt,
  partopsep=0pt,
  leftmargin=*
}

\usepackage[preprint]{acl}

\usepackage{times}
\usepackage{latexsym}

\usepackage[T1]{fontenc}

\usepackage[utf8]{inputenc}

\usepackage{microtype}

\usepackage{inconsolata}

\usepackage{graphicx}

\title{Modeling Discourse Profiles in Climate Misinformation Discourse}

\author{Ge Gao, Zhengyang Shan, Bryan A. Plummer\textsuperscript{1}, Margrit Betke, Derry Wijaya \\
        Boston University \\ \{ggao02, shanzy, bplum, betke, wijaya\}@bu.edu}

\begin{document}
\maketitle
\begin{abstract}
Prior work on climate misinformation detection typically classifies individual comments or claims using text alone. This framing overlooks a key property of online climate discourse: misinformation is often produced, reinforced, and sustained through recurring rhetorical patterns that reappear across comments and threads. Using a large-scale TikTok climate discourse dataset, we construct discourse profiles by clustering climate-related comments into recurring rhetorical patterns. We then incorporate these profiles into fine-grained climate misinformation classifiers using a late-fusion architecture.  Under video-disjoint evaluation across three training seeds, discourse-profile-aware models significantly improve macro-F1 over text-only baselines for bag-of-words, TF-IDF, and MPNet representations. RoBERTa-large exhibits a positive mean improvement but substantially greater variability, and the difference is not statistically significant. Beyond classification, our results show discourse profiles shape misinformation reception and propagation. Discourse-profile-conditioned language model agents exhibit misinformation acceptance patterns aligned with profile-level misinformation prevalence in \textsc{ClimateDisc}, and the relationship between misinformation prevalence and engagement persistence differs across dominant discourse profiles.
\end{abstract}

\footnotetext[1]{Work done while at BU}

\section{Introduction}
\begin{figure}[t]
\centering
\includegraphics[width=1\columnwidth]{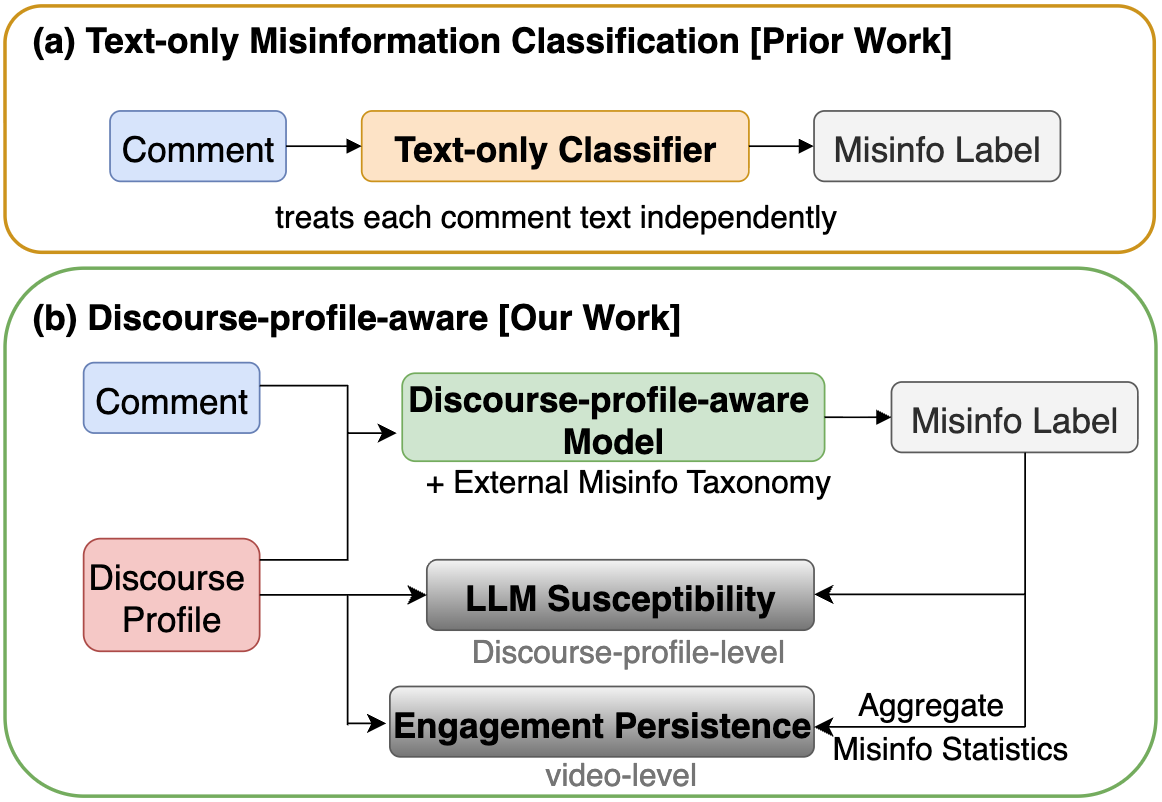}
\caption{Comparison of text-only vs.\ discourse-profile-aware misinformation modeling. (a) Prior work \cite{priorClimate1, priorClimate2, priorClimate3, priorClimate4, priorClimate5} typically predicts a misinformation category from comment text alone, treating each comment independently. (b) We construct discourse profiles by clustering comments and use discourse profile as additional context for misinformation-type prediction; the same discourse profile abstraction also supports discourse-profile-conditioned analyses of LLM susceptibility and engagement persistence.}
\vspace{-0.4em}
\label{fig:overview}
\end{figure}

Short-video social media platforms such as TikTok have become major venues for public discussion of climate change, where accurate information, skepticism, and misinformation coexist in the same high-volume comment threads \cite{climateOnSocial1, climateOnSocial2, climateOnTK1, climateOnTK2, tkMisinfo1}. This setting is particularly challenging for automated analysis: short, algorithmically amplified videos drive rapid diffusion, while discussions are fast-moving, emotionally charged, allowing misinformation to persist and resist correction.

As shown in Figure~\ref{fig:overview}a, most prior work uses text-only classifiers that label each comment or claim in isolation \cite{priorClimate1, priorClimate2, priorClimate3, priorClimate4, priorClimate5}. However, climate discourse also exhibits recurring narrative frames and rhetorical styles that reappear across threads. This suggests that \emph{how people consistently argue} can provide complementary signal beyond the words in a single comment.

We introduce \emph{discourse profiles} to address this challenge: a compact, data-driven abstraction of recurring narrative frames and rhetorical styles in climate discourse.
This perspective complements prior work showing that incorporating speaker and social context can improve misinformation detection beyond text alone \cite{nguyen2020fang, mehta2022gnn}, and that beliefs and framing are not fully captured by lexical features \cite{allaway2021zero, zhang2023doublehtwitteruserstance}. Rather than relying on predefined demographics or static author embeddings, we construct discourse profiles directly from large-scale, in-the-wild TikTok comments.
Specifically, we cluster semantic comment embeddings and use LLM-based cluster consolidation to group the resulting fine-grained clusters into a small set of interpretable discourse profiles. Importantly, profiles are assigned at the \emph{comment level}, not the user level, allowing the same user to exhibit different discourse patterns across contexts. 

We then use discourse profiles in three ways (Figure~\ref{fig:overview}b):
(i) to augment fine-grained misinformation-type prediction,
(ii) to construct a controlled susceptibility probe using discourse-profile-conditioned LLM agents, and
(iii) to analyze how misinformation relates to engagement persistence across different discourse contexts. 

Across video-disjoint evaluation with three training seeds, discourse-profile conditioning significantly improves macro-F1 for bag-of-words, TF-IDF, and MPNet classifiers, while RoBERTa-large shows a positive but statistically inconclusive mean difference. Discourse-profile-conditioned LLM agents also exhibit susceptibility patterns that align with profile-level misinformation prevalence in CLIMATEDISC. Finally, the relationship between misinformation prevalence and engagement persistence varies across dominant discourse profiles.

Our main contributions are:
\begin{itemize}
    \setlength{\itemsep}{2pt}
    \setlength{\parskip}{0pt}
    \setlength{\parsep}{0pt}
    \item We introduce a data-driven discourse-profile abstraction for climate misinformation discourse, constructed from semantic clustering of large-scale TikTok comments and consolidation into interpretable profile types.
    \item We develop a discourse-profile-aware modeling and evaluation framework that incorporates profile information into fine-grained misinformation classification and uses matched text-only, shuffled-profile, and stylistic-variation controls to isolate when profile information contributes beyond text.
    \item We extend the same discourse-profile abstraction to controlled LLM susceptibility probing and video-level temporal analysis, revealing that the association between misinformation prevalence and engagement persistence differs across discourse profiles.
\end{itemize}


\section{Related Work}

\subsection{Climate Misinformation Narratives}

Climate misinformation includes not only outright denial, but also broader obstruction narratives such as questioning climate solutions, emphasizing economic trade-offs, or attacking scientific credibility \cite{lamb, priorClimate2, holder}. We adopt the fine-grained \textsc{QuotaClimat} taxonomy \cite{QuotaClimat}, which captures diverse climate obstruction narratives (Appendix~\ref{appendix:disinfo}).
Prior climate-misinformation and stance-detection methods discussed above typically predict labels from individual textual instances \cite{priorClimate2, priorClimate4, priorClimate3, priorClimate5}. In contrast, we model recurring rhetorical patterns through discourse profiles and analyze how different profiles preferentially express specific misinformation types.

\subsection{Author Context and Persona in NLP}

Prior work shows that author context can improve stance and misinformation detection by capturing latent beliefs, interaction patterns, and communicative styles beyond text alone \cite{benton-dredze-2018-using, Qiu2015ModelingUA, lynn-etal-2019-tweet, barel-etal-2025-acquired}. Related work has also explored dynamic and multi-persona user representations \cite{del-tredici-etal-2019-shall, marsden2019}, while studies of climate denial rhetoric show that misinformation discourse often relies on recurring rhetorical and networking strategies rather than isolated claims \cite{article}. Existing approaches typically rely on predefined attributes, interaction structures, or static user embeddings. We instead construct discourse profiles directly from large-scale climate discourse and use them both for prediction and downstream behavioral analysis.

\subsection{LLM Profiles and Engagement Dynamics}

Recent work conditions LLMs on personas or discourse profiles to simulate subgroup behavior \cite{llm1, llm2, llm3, llmpersona}. We instead use discourse-profile-conditioned LLM agents as a controlled probe for misinformation susceptibility.
Prior work has examined misinformation diffusion and persistence through content, emotion, and network structure \cite{Vosoughi2018, liu-etal-2023-analyzing, Grinberg2019, SUN2024108271}. We extend this line of work by showing that engagement persistence varies across discourse profiles.

\begin{figure*}[t]
  \centering
  \includegraphics[width=0.70\linewidth]{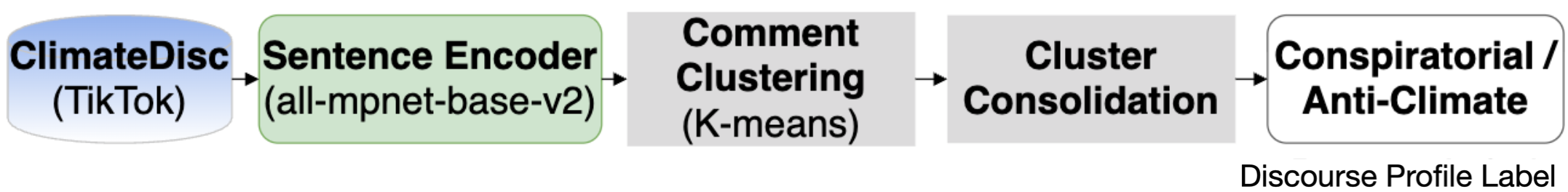}
  \caption{Offline discourse profile construction: encode comments, cluster embeddings, and use GPT-5-based cluster consolidation to assign the resulting clusters to higher-level discourse profiles $d_i$ (Section \ref{sec:persona_creation}).}
  \label{fig:archia}
\vspace{-0.4em}
\end{figure*}

\begin{figure*}[t]
  \centering
  \includegraphics[width=0.9\linewidth]{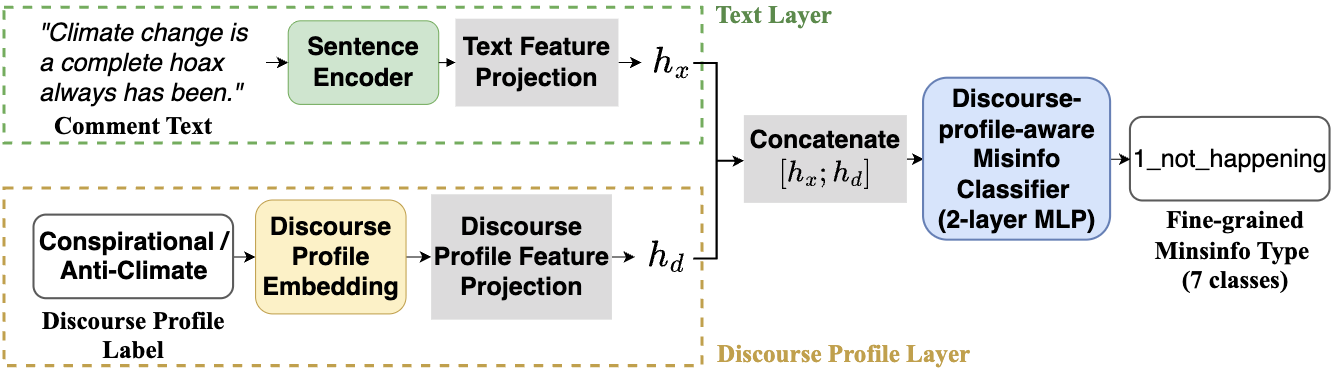}
  \caption{Late-fusion discourse-profile-aware misinformation modeling: obtain misinformation-type labels and evaluate prediction models that concatenate contextual text representations with discourse-profile embeddings before classification (Section~\ref{sec:persona_enhance_pred}). Details of how we obtain comment-level misinformation labels are in Appendix~\ref{appendix:taxonomy_transfer}.}
  
  \label{fig:archib}
\vspace{-0.4em}
\end{figure*}

\section{Method}

Given short-form social media comments discussing climate change, our goal is to characterize and predict fine-grained climate misinformation by explicitly modeling recurring discourse-level regularities. For each comment $c_i$, we (i) assign a discourse profile $d_i$ capturing recurring rhetorical patterns, (ii) obtain a fine-grained misinformation type $y_i$ when applicable, and (iii) analyze how discourse profile conditions misinformation susceptibility and engagement persistence at both the comment and video levels. 

Our pipeline proceeds in two stages. First, we construct discourse profiles from large-scale climate comment discourse (Figure~\ref{fig:archia}; Section~\ref{sec:persona_creation}). Second, we incorporate discourse profile as an explicit signal for misinformation-type prediction (Figure~\ref{fig:archib}) and for downstream analyses of susceptibility and engagement. Comment-level misinformation labels are obtained using a classifier trained on the curated \textsc{QuotaClimat} taxonomy. Because this requires cross-domain transfer from curated claims to in-the-wild TikTok comments, we evaluate label quality using claim--comment matching, an independent LLM comparison, and a two-annotator human validation study on 100 randomly sampled \textsc{ClimateDisc} comments (Appendix~\ref{appendix:taxonomy_transfer}).

\subsection{Discourse Profile Creation}
\label{sec:persona_creation}

To capture recurring rhetorical and discourse patterns used in Sections \ref{sec:persona_enhance_pred}-\ref{sec:temporalDynamics}, we construct discourse profiles from large-scale climate change comments in \textsc{ClimateDisc} \cite{gao-etal-2025-insights}, a large-scale corpus of TikTok climate change videos and associated comments (Section \ref{sec:datasets}), with the resulting profiles serving as a discourse-level abstraction across downstream tasks.

We represent comments as dense semantic embeddings using a pre-trained MPNet-based sentence transformer \cite{all-mpnet-base-v2}. We then cluster the embeddings using $k$-means \cite{kmeans}, selecting the number of clusters by jointly examining Within-Cluster Sum of Squares (WCSS) and Silhouette scores over a candidate range of $k \in [3, 30]$. This analysis identifies $k=21$ as a practical trade-off between cluster compactness and separation (Section~\ref{sec:results-personas}). We additionally compared against HDBSCAN on the same embeddings (Appendix \ref{app:hdbscan}). 

To obtain a compact discourse-level abstraction for downstream modeling, we use GPT-5 for cluster consolidation. We prompt the model to group the 21 fine-grained semantic clusters into higher-level \emph{discourse profiles} according to shared rhetorical patterns and communicative stances toward climate change (Appendix \ref{app:cluster_consolidation}). This procedure yields six higher-level discourse profiles, along with a residual ‘Noise / Low-Signal’ group excluded from discourse-profile-sensitive analyses. The resulting discourse profile definitions are reported in Table~\ref{tab:personas}.
\begin{enumerate}[leftmargin=1.5em, itemsep=0pt, topsep=2pt, parsep=0pt]
    \item \textit{``Pennsylvania, same dread.''}
    \item \textit{``We're having an unseasonably warm winter here in Minnesota.''}
    \item \textit{``This winter in Michigan there was only snow for a few weeks... now it's 50 degrees and sunny.''}
\end{enumerate}
While these comments differ in surface form, they all ground climate discussion in place-based lived experience. We therefore treat them as reflecting a shared discourse-level profile.

With discourse profiles defined, we next examine whether this abstraction provides predictive signal beyond comment text alone. Specifically, we evaluate whether incorporating discourse profile information improves fine-grained climate misinformation classification.

\begin{table}[t]
\centering
\scriptsize
\setlength{\tabcolsep}{3pt}
\renewcommand{\arraystretch}{1.05}
\begin{tabular}{|@{\hspace{2pt}}p{0.49\linewidth}@{}|@{\hspace{2pt}}p{0.49\linewidth}@{}|}
\hline
\textbf{Local / Experiential Observers} &
\textbf{Ideological / Political Combatants} \\
1. Local Weather Anecdotes &
0. Anti-Trump/GOP/Biden \\
12. Storms \& Flooding &
11. Government Blame / Anarchy \\
13. Geographical Self-Mentions &
18. Electoral Mobilization \\
\hline
\textbf{Systemic / Structural Critics} &
\textbf{Pro-Science / Rationalist} \\
4. Consumption / Plastics Skepticism &
14. Pro-Science / Anti-Misinformation \\
17. Energy / Policy Debate & \\
19. Agriculture / Diet & \\
20. Anti-Corporation / Capitalism & \\
\hline
\textbf{Doomism / Nihilism} &
\textbf{Conspiratorial / Anti-Climate} \\
8. Doomism &
7. Natural causes – not man-made \\
& 16. Denial / Hoax / Manipulation \\
\hline
\multicolumn{2}{|@{}p{\linewidth}@{}|}{%
\hspace{2pt}\textbf{Noise / Low-Signal (Excluded)}:
\hspace{2pt}2. Emoji / Applause Spam;
\hspace{2pt}3. Minimal Agreement Tokens;
\hspace{2pt}5. Polite Thanks;
\hspace{2pt}6. Short Emotional Bursts;
\hspace{2pt}9. Excited Agreement;
\hspace{2pt}10. Shared Anxiety;
\hspace{2pt}15. @ Mentions.%
} \\
\hline
\end{tabular}
\vspace{-0.4em}
\caption{Discourse profiles obtained by grouping the 21 $k$-means clusters of comment embeddings into semantically coherent discourse-profiles. The Noise / Low-Signal group is excluded from downstream discourse-profile-sensitive analyses.}
\label{tab:personas}
\vspace{-0.4em}
\end{table}

\subsection{Profile-Aware Misinfo Classification}
\label{sec:persona_enhance_pred}

We compare text-only and discourse-profile-aware classifiers to test whether discourse profile provides predictive signal beyond comment text. Our goal is not to maximize benchmark performance, but to isolate when discourse profile contributes complementary information under different text representations and model capacities.

\paragraph{Text representations.}
Given a comment \(t\) in \textsc{ClimateDisc}, we encode it as a feature vector \(\mathbf{x}=\phi(t)\), where \(\phi(\cdot)\) may denote either a pretrained dense sentence encoder or a lexical featurizer. In particular, we consider three families of text representations: (i) MPNet-based sentence embeddings \cite{all-mpnet-base-v2}, (ii) non-contextual bag-of-words features, and (iii) TF-IDF \cite{tfidf} features. This allows us to examine the effect of discourse profile across both dense semantic and sparse lexical input spaces.

\paragraph{Text-only baseline.}
Model~A is a text-only classifier that predicts fine-grained misinformation type from \(\mathbf{x}\) alone. The text representation is passed through a nonlinear projection \(\mathbf{h}_x=f_x(\mathbf{x})\), where \(f_x(\cdot)\) denotes an affine transformation followed by ReLU and dropout, and then mapped to output logits \(\hat{\mathbf{y}}=g(\mathbf{h}_x)\). The prediction space follows the seven-class misinformation taxonomy from \textsc{QuotaClimat}, excluding \(0\_\texttt{not\_relevant}\).

\paragraph{Discourse-profile-aware model.}
We incorporate discourse profile into misinformation prediction. Each discourse profile label \(c \in \{1,\ldots,6\}\) is mapped to a learnable embedding \(\mathbf{d}=E(c)\).

We extend the text-only baseline by incorporating discourse profile as an additional latent signal. The discourse-profile embedding is projected as \(\mathbf{h}_d=f_d(\mathbf{d})\), then concatenated with the text representation \(\mathbf{h}_x\) to form \(\mathbf{h}=[\mathbf{h}_x;\mathbf{h}_d]\). A joint nonlinear transformation \(\mathbf{z}=f_j(\mathbf{h})\) is then applied before the final classifier, yielding \(\hat{\mathbf{y}}=g(\mathbf{z})\). This formulation introduces discourse profile only at the classifier stage while keeping the contextual text representation unchanged.

\paragraph{Multi-seed evaluation.} We use a fixed video-disjoint train/validation/test partition and repeat model training using three random seeds. We report accuracy and macro-F1 as mean across these runs. To assess whether correct-profile conditioning improves over the corresponding text-only model, we conduct a two-sided paired t-test on the three seed-level macro-F1 values.

\paragraph{Isolating profile signal.}
To distinguish whether improvements arise from discourse-profile information or merely increased model capacity, we conduct a controlled profile permutation test in which we randomly shuffle profile labels while holding the model architecture and training procedure fixed. For each of the three training seeds, we train ten shuffled-profile models using independently permuted profile assignments. We compare the shuffled-profile macro-F1 distribution with the corresponding correct-profile score using a two-sided one-sample t-test.

\paragraph{Robustness under stylistic variation.}
We additionally evaluate robustness under LLM-generated stylistic rewrites. Following \citet{mpcg}, we generate multi-round discourse-profile-conditioned variants that preserve the underlying misinformation category while varying rhetorical style and surface realization. We then evaluate both text-only and discourse-profile-aware classifiers on the rewritten comments. Generation details, prompts, and examples are provided in Appendix~\ref{app:llm_variants_section}.

\subsection{Discourse-Profile-Level Misinformation Susceptibility with LLM}
\label{sec:suspectiblescore}

We use discourse-profile-conditioned LLM agents as a controlled probe of whether the constructed discourse profiles recover systematic differences in misinformation susceptibility. We construct a balanced evaluation set of 350 misinformation claims from \textsc{QuotaClimat}, sampling 50 claims from each of the seven categories. These claims are separate from the \textsc{ClimateDisc} comments used to construct the discourse profiles. We use GPT-5 as the language model for the discourse-profile-conditioned susceptibility probe. For each inferred discourse profile, we construct a natural-language profile card summarizing characteristic beliefs, goals, tone, and rhetorical tendencies (Appendix~\ref{app:persona_card_construction}). The cards do not contain profile-level misinformation rates, category-specific acceptance rates, or an expected ordering of profiles by susceptibility. Because the cards nevertheless encode characteristic beliefs and rhetorical tendencies, we interpret this experiment as a controlled profile-specific probe rather than a simulation of individual users.
 
As a specificity control, we randomly shuffle the profile-to-card mapping 100 times while keeping the claim set and evaluation procedure fixed. For each permutation, we recompute the correlation between the \textsc{ClimateDisc} profile-level misinformation rates and LLM susceptibility scores.

For discourse profile $d$, we define an LLM susceptibility score as the proportion of misinformation claims not rejected by the discourse-profile-conditioned LLM agent: 
{
\setlength{\abovedisplayskip}{3pt}
\setlength{\belowdisplayskip}{3pt}
$$
L_d = \frac{1}{|Q|} \sum_{q \in Q} \mathbb{I}[r_{d,q} \neq \mathrm{reject}],
$$
}
where $Q$ is the set of evaluation claims, $r_{d,q}$ is the LLM's response for discourse profile $d$ on claim $q$, and $\mathbb{I}[\cdot]$ denotes an indicator function.

\paragraph{Human--LLM Discourse Profile Alignment.}
\label{sec:align}
We assess whether discourse-profile-conditioned LLM susceptibility patterns align with those of real human discourse profiles. 
Let the comments in \textsc{ClimateDisc} be indexed by \(i \in \{1,\ldots,N\}\), where \(d_i\) denotes the discourse profile assigned to comment \(i\) and \(y_i\) denotes its misinformation label. For each discourse profile \(d\), we define the set of comments assigned to that profile as
\[
C_d = \{i \in \{1,\ldots,N\} : d_i = d\}.
\]
We then define the human misinformation rate for profile \(d\) as
\[
H_d =
\frac{1}{|C_d|}
\sum_{i \in C_d}
\mathbb{I}[y_i \neq 0],
\]
where \(y_i = 0\) denotes \texttt{not\_relevant} and \(y_i \in \{1,\ldots,7\}\) denotes one of the seven misinformation categories. Thus, \(H_d\) represents the proportion of comments associated with discourse profile \(d\) that are labeled as misinformation.
We compute the LLM susceptibility score $L_d$ previously defined.
We then measure alignment between $(H_d)$ and $(L_d)$ using Pearson and Spearman correlations, where higher correlation indicates closer correspondence between human and LLM discourse profile behavior. We next examine whether these discourse-profile-level differences translate into distinct engagement trajectories.

\subsection{Discourse-Profile Temporal Dynamics}
\label{sec:temporalDynamics}

To examine how discourse profile conditions the downstream impact of misinformation, we analyze engagement persistence at the video level. For each video in \textsc{ClimateDisc} we compute (i) its dominant discourse profile, defined as the majority profile among its comments, (ii) its misinformation rate, and (iii) engagement persistence, measured as the time required to reach 60\% of total comments \cite{liu-etal-2023-analyzing}. Our dataset for this engagement persistence experiment contains 2,925 videos\footnote{We exclude videos with zero or one comment, since persistence requires at least two comments to be defined.}.

Formally, for a video $v$ with $N_v$ total comments, let $t_{v,i}$ denote the timestamp of the $i$-th comment. We define the cumulative number of comments received by time $t$ as
{
\setlength{\abovedisplayskip}{3pt}
\setlength{\belowdisplayskip}{3pt}
$$
S_v(t) = \sum_{i=1}^{N_v} \mathbb{I}[t_{v,i} \le t],
$$
} 
where $\mathbb{I}[\cdot]$ is the indicator function. Let $t_v^{(0)}$ denote the timestamp of the first observed comment for video $v$. The time at which the video reaches a fraction $\alpha$ of its total comments is defined as $T_v(\alpha) = \inf \{ t : S_v(t) \ge \alpha N_v \}.$
We define engagement persistence as
$
P_v = T_v(0.6) - t_v^{(0)},
$
which measures the duration (in hours) required for video $v$ to accumulate 60\% of its total comments. Larger values of $P_v$ indicate more sustained engagement over time.

To examine whether persistence patterns differ across discourse contexts, we stratify videos into low, mid, and high misinformation groups by misinformation rate. Within each discourse profile type, we conduct Kruskal--Wallis tests to assess whether persistence distributions differ significantly across misinformation levels. 

\begin{table*}[t]
\centering
\small
\begin{tabular}{lccccccc}
\toprule
\textbf{Representation} & \textbf{Text Acc} & \textbf{Prof Acc} & \textbf{$\Delta$ Acc} & \textbf{Text F1} & \textbf{Prof F1} & \textbf{$\Delta$ F1} & \textbf{Paired $p$} \\
\midrule
\multicolumn{8}{l}{\textit{[MLP]}} \\
Bag-of-Words
& 0.690 & 0.709 & +0.019
& 0.586 & 0.616 & +0.030
& 0.028 \\
TF-IDF
& 0.675 & 0.709 & +0.034
& 0.522 & 0.601 & +0.080
& 0.006 \\
MPNet
& 0.799 & 0.808 & +0.010
& 0.726 & 0.739 & +0.014
& 0.015 \\
\midrule
RoBERTa-large
& 0.845 & 0.856 & +0.011
& 0.794 & 0.810 & +0.015
& 0.381 \\
\bottomrule
\end{tabular}
\caption{Comparison of text-only (\textbf{Text}) and discourse-profile-aware (\textbf{Prof}) misinformation classification under video-disjoint evaluation. Accuracy and macro-F1 values are means across three training seeds. $p$-values are from two-sided paired t-tests comparing seed-level macro-F1 scores between the correct-profile and corresponding text-only models. Overall, correct-profile conditioning improves macro-F1 for bag-of-words, TF-IDF, and MPNet, while the RoBERTa-large difference is not statistically significant across seeds. }\label{tab:main_results}
\end{table*}

\section{Experiments}


\subsection{Datasets}
\label{sec:datasets}
\paragraph{\textsc{QuotaClimat.}}
To obtain fine-grained misinformation types, we use \textsc{QuotaClimat} \cite{QuotaClimat}, a curated dataset of 6{,}091 climate-related statements annotated with one of the misinformation categories specified in Appendix~\ref{appendix:disinfo}. We use \textsc{QuotaClimat} for two purposes:
(i) training a misinformation classifier that is transferred to \textsc{ClimateDisc} comments (Appendix~\ref{appendix:taxonomy_transfer}),
and (ii) susceptibility probing with discourse-profile-conditioned LLM agents using a balanced set of 350 claims (50 per category; Section~\ref{sec:suspectiblescore}).

\paragraph{\textsc{ClimateDisc.}}
For discourse profile creation and downstream in-the-wild analyses, we rely on the \textsc{ClimateDisc} dataset \cite{gao-etal-2025-insights}, which comprises transcribed English speech from 7{,}110 TikTok videos and 116{,}256 corresponding comments tagged with ``\#climatechange,'' collected between January~2024 and November~2024. We use \textsc{ClimateDisc} in:
(i) discourse profile creation (Section~\ref{sec:persona_creation}),
(ii) discourse-profile-aware misinformation classification on in-the-wild comments (Section~\ref{sec:persona_enhance_pred}),
(iii) human discourse-profile-level misinformation rates $H_d$ for LLM alignment (Section~\ref{sec:align}),
and (iv) temporal dynamics and engagement persistence (Section~\ref{sec:temporalDynamics}).

For the discourse-profile-aware misinformation classification experiments on \textsc{ClimateDisc} (Section~\ref{sec:persona_enhance_pred}), comments are divided into training, validation, and test sets in a 0.8/0.1/0.1 ratio. The splits are video-disjoint. All comments from a given video occur in exactly one partition, preventing video-level overlap between training and evaluation data. The resulting split contains 2,717 training videos, 344 validation videos, and 347 test videos. Noise / Low-Signal comments are excluded.

\subsection{Implementation Details}

For discourse-profile-aware misinformation classification, we use lightweight feed-forward models with MPNet embeddings, bag-of-words features, and TF-IDF features as input. For higher-capacity contextual encoders, we instantiate text-only and discourse-profile-aware variants using RoBERTa-large \cite{roberta}. We use the fixed video-disjoint train/validation/test split described above. Lightweight models are trained with Adam using learning rate $3 \times 10^{-4}$, batch size 32, and early stopping on validation macro-F1. RoBERTa-large models are trained with Adam using learning rate $2 \times 10^{-5}$, batch size 8, maximum sequence length 128, and early stopping on validation macro-F1. All models use cross-entropy loss and gradient clipping at 1.0.

\subsection{Results}
\label{sec:results}


Table~\ref{tab:main_results} summarizes misinformation classification performance under the video-disjoint evaluation setting. All reported accuracy and macro-F1 values are averaged across three training seeds. For the lightweight MLP-based classifiers, incorporating the correct discourse profile improves performance across all three text representations.

For bag-of-words features, discourse-profile conditioning increases mean accuracy from 0.690 to 0.709 and macro-F1 from 0.586 to 0.616, corresponding to a macro-F1 improvement of +0.030. The improvement is significant across training seeds under a paired $t$-test ($p=0.028$). The largest gain is observed with TF-IDF features, where mean accuracy increases from 0.675 to 0.709 and macro-F1 from 0.522 to 0.601 ($\Delta$F1 = +0.080, $p=0.006$). With frozen MPNet sentence embeddings \cite{Song2020MPNetMA}, the profile-aware model improves mean accuracy from 0.799 to 0.808 and macro-F1 from 0.726 to 0.739 ($\Delta$F1 = +0.014, $p=0.015$). These results indicate that explicit discourse-profile information provides complementary predictive signal
across both sparse lexical and dense sentence-level representations.

The effect is less conclusive for the higher-capacity contextual encoder. For RoBERTa-large, the profile-aware model achieves higher mean accuracy (0.856 vs.\ 0.845) and macro-F1 (0.810 vs.\ 0.794), yielding an average macro-F1 difference of +0.015. However, this difference varies substantially across training seeds and is not statistically significant ($p=0.381$). We therefore interpret the RoBERTa-large result as a positive but statistically inconclusive average difference rather than evidence of a reliable improvement from discourse-profile conditioning.

To examine whether the aggregate gains are concentrated in only a small subset of misinformation types, we additionally conduct a category-level error analysis in Appendix~\ref{app:error_analysis}.

\paragraph{Contribution of Correct Profile Assignment.}
To isolate discourse-profile signal from added model capacity, we compare correct-profile conditioning against shuffled-profile controls averaged over 10 random permutations (Table~\ref{tab:persona_permutation}). 
The shuffled-profile control reveals substantial differences across text representations. For bag-of-words, correct profiles outperform shuffled profiles in all three training seeds, with a mean macro-F1 advantage of \(+0.034\); shuffled-profile performance is approximately equal to the text-only baseline. TF-IDF likewise shows a consistent correct-versus-shuffled advantage of \(+0.040\) across all three seeds. These results provide the clearest evidence that the specific comment–profile correspondence contributes predictive information in the sparse lexical settings.
The pattern is weaker for dense representations. For MPNet, correct profiles outperform shuffled profiles by only \(+0.003\) macro-F1 on average, and the correct-versus-shuffled contrast is significant in only one of three seeds. Thus, although the correct-profile MPNet model significantly exceeds the text-only model across seeds, much of that improvement cannot be uniquely attributed to the semantic correctness of the profile assignment. RoBERTa-large similarly exhibits unstable shuffled-profile behavior, with a significant correct-versus-shuffled difference in only one of three seeds.


\subsubsection{Discourse Profile Construction}
\label{sec:results-personas}
\begin{figure}[t]
\centering
\includegraphics[width=0.9\columnwidth]{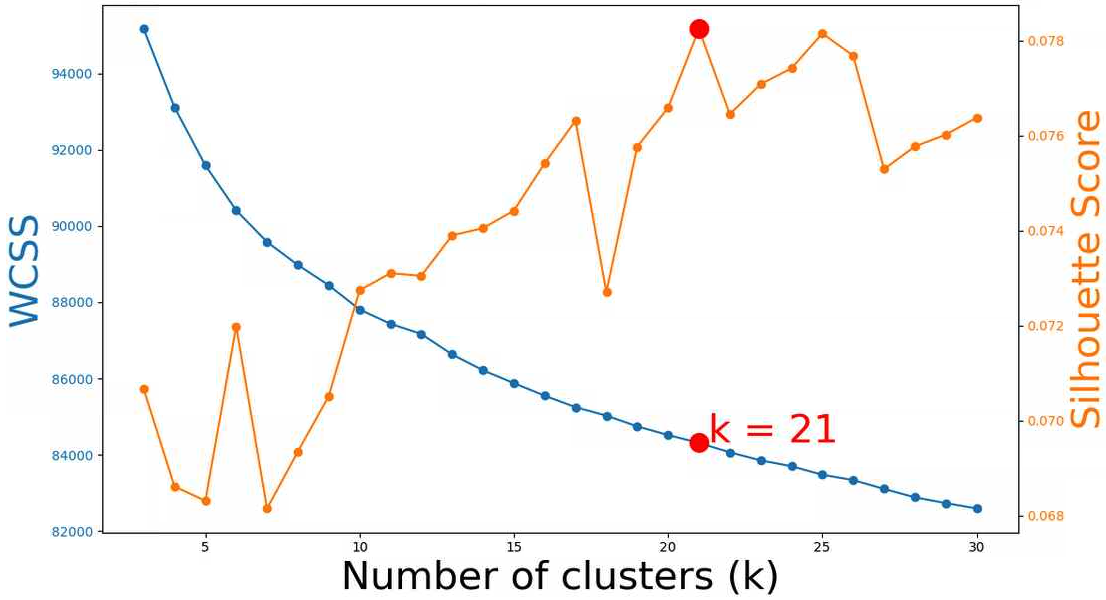} 
\caption{Within-Cluster Sum of Squares (WCSS; lower is better) and Silhouette Score (higher is better) across varying numbers of clusters $k$. 
The Silhouette score reaches a local maximum at \(k=21\).
}
\label{wcss}
\vspace{-0.4em}
\end{figure}

As described in Section \ref{sec:persona_creation}, we construct the initial discourse profiles using k-means clustering and select the number of clusters \(k\) by jointly examining WCSS and Silhouette scores over \(k \in [3,30]\). 
Figure~\ref{wcss} reports these cluster-selection diagnostics.
WCSS decreases smoothly without a decisive elbow, whereas the Silhouette score reaches a local maximum at \(k=21\). We therefore set $k=21$ as a practical trade-off between separation and interpretability. As a qualitative check, Appendix~\ref{appendix:samples} lists the ten comments nearest each centroid, which exhibit consistent topical and rhetorical patterns.

We aggregate the 21 clusters into six higher-level discourse profiles plus a residual ``Noise / Low-Signal'' group. After consolidation, the six higher-level discourse profiles remain relatively balanced (Appendix ~\ref{app:cluster_dist}), reducing the likelihood that downstream effects are driven by a single dominant profile.
We exclude the Noise group from discourse-profile-sensitive analyses (it consists largely of short acknowledgments, emojis, and @-mentions).
Table~\ref{tab:personas} summarizes the mapping.

Across ten independent k-means initializations at \(k=21\), mean pairwise adjusted Rand index is \(0.741\) and mean normalized mutual information is \(0.815\). Cluster-size coefficients of variation range from \(0.301\) to \(0.448\), and after consolidation into the higher-level discourse profiles, average profile-assignment agreement remains \(82.8\%\).
As a supplemental interpretability check, we asked two independent LLM raters, GPT-5.5 and Claude Opus 4.8, to reassign each of the 21 clusters. The raters were blind to the original cluster-to-profile mapping and assigned each cluster to one of the six discourse profiles or Noise / Low-Signal. The two raters agreed on 19 of 21 clusters (\(\kappa = 0.88\)). Their agreement with the original consolidation was \(\kappa = 0.71\) and \(0.64\), respectively (Appendix~\ref{app:cluster_validation}).

\begin{table*}[t]
\centering
\small
\begin{tabular}{lccccc}
\toprule
\textbf{Representation} & \textbf{Text F1} & \textbf{Shuffled F1} & \textbf{Correct F1} & \textbf{Correct $-$ Shuffled} & \textbf{Significant seeds} \\
\midrule
\multicolumn{6}{l}{\textit{[MLP]}} \\
Bag-of-Words
& 0.586 & 0.582 & 0.616 & +0.034 & 3 / 3 \\
TF-IDF
& 0.522 & 0.561 & 0.601 & +0.041 & 3 / 3 \\
MPNet
& 0.726 & 0.736 & 0.739 & +0.003 & 1 / 3 \\
\midrule
RoBERTa-large
& 0.794 & 0.776 & 0.810 & +0.033 & 1 / 3 \\
\bottomrule
\end{tabular}
\caption{Permutation-test evaluation of discourse-profile conditioning across text representations. Shuffled-profile scores are first averaged across ten independent profile shuffles within each training seed; reported values are then averaged across the three training seeds. The last column reports how many of the three random seeds show a significant correct-vs-shuffled difference under the one-sample t-test. The correct-versus-shuffled advantage is most consistent for bag-of-words and TF-IDF, indicating that the specific profile assignment contributes most clearly in sparse lexical settings.}
\label{tab:persona_permutation}
\end{table*}

\subsubsection{Discourse-profile-Level Susceptibility Probing with LLM}

\paragraph{Human vs. LLM Agents Susceptibility.}
We observe a strong positive correlation between human and LLM discourse profile susceptibility patterns. The Pearson correlation between $H_d$ and $L_d$ (specified in Section \ref{sec:align}) is 0.82, indicating substantial linear agreement. The Spearman rank correlation is even higher at 0.94, suggesting that the relative ordering of discourse profiles by misinformation susceptibility is highly consistent between humans and LLM agents. This alignment suggests that discourse-profile-conditioned LLM agents capture meaningful, discourse-profile-specific structure in misinformation acceptance.

\paragraph{Profile card permutation control.}
To test whether this alignment depends on the correct profile card correspondence, we randomly permute the profile-to-card mapping 100 times while keeping the claim set and evaluation procedure fixed. Under permutation, the mean Spearman correlation is $\rho=-0.10$, whereas the true profile card assignment yields $\rho=0.94$ and is significant under the permutation test ($p=0.03$). This indicates that the observed correlation depends on the specific profile card correspondence rather than only on a generic tendency of the LLM to accept or reject the claims.

\paragraph{Susceptibilities Across Discourse Profiles.}
Table~\ref{tab:overall_susceptibility} summarizes overall susceptibility score (specified in Section \ref{sec:suspectiblescore}) per discourse profile.
\textit{Conspiratorial / Anti-Climate} shows very high susceptibility (96.3\%), while other discourse profiles range from 4.9\% to 29.4\%. Notably, \textit{Ideological / Political Combatants} show the lowest overall susceptibility (4.9\%), suggesting that strong political engagement does not necessarily translate to greater vulnerability to misinformation. Similarly, the \textit{Pro-Science / Rationalist}  demonstrates comparatively low susceptibility (10.4\%), though this group is not immune to misinformation, indicating that scientific self-identification alone does not guarantee resistance. \textit{Doomism / Nihilism} is comparatively higher (22.0\%), highlighting the role of affective orientations such as resignation or fatalism in shaping misinformation vulnerability. 

\begin{table}[t]
\centering
\small
\begin{tabular}{lc}
\hline
\textbf{Discourse Profile} & \textbf{Susceptible (\%)} \\
\hline
Local / Experiential Observers & {29.4} \\
Ideological / Political Combatants & {4.9} \\
Systemic / Structural Critics & {8.0} \\
Pro-Science / Rationalist & {10.4} \\
Doomism / Nihilism & {22.0} \\
Conspiratorial / Anti-Climate & \textbf{96.3} \\
\hline
\end{tabular}
\caption{Overall misinformation susceptibility rate by discourse profile. Conspiratorial / Anti-Climate profile exhibits the highest acceptance rate.}
\vspace{-0.4em}
\label{tab:overall_susceptibility}
\end{table}

\subsubsection{Engagement Dynamics Across Profiles}




\begin{figure}[t]
\centering
\vspace{-0.4em}
\includegraphics[width=1\columnwidth]{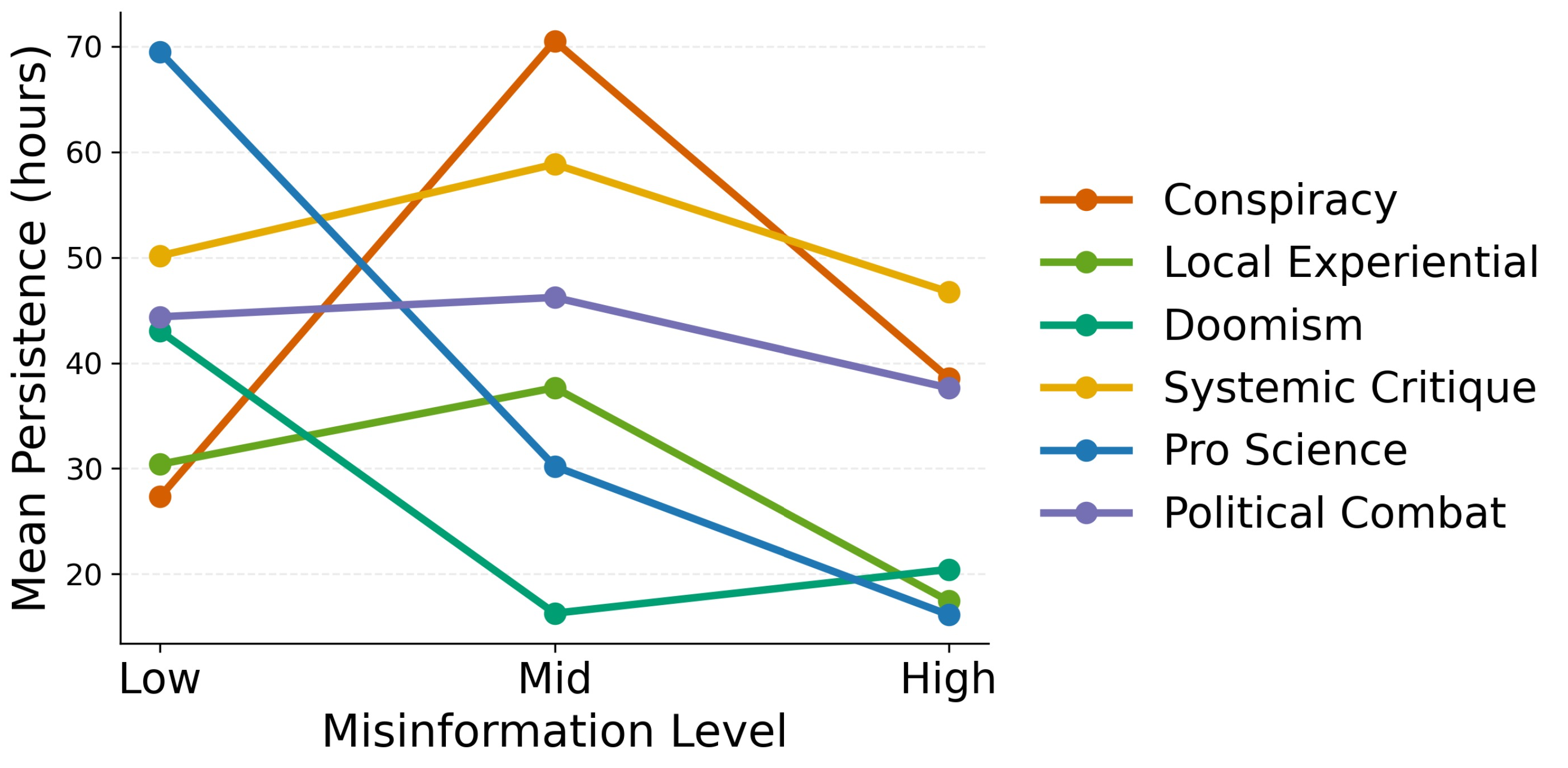} 
\caption{Mean persistence hours across misinformation level (Low, Mid, High) for each profile type.}
\label{interaction_lineplot}
\vspace{-0.4em}
\end{figure}

To examine how persistence co-varies with misinformation prevalence, we group videos into Low, Mid, and High misinformation bins and compute mean persistence hours for each profile-bin combination (Figure \ref{interaction_lineplot}). For \textit{Pro Science}, persistence is higher in the low-misinformation group (69.5 hours) and lower in the high-misinformation group (16.1 hours). Within the \textit{Conspiracy} profile, the high-misinformation group exhibits higher mean persistence than the low-misinformation group. \textit{Systemic Critique} exhibits relatively small variation across bins. \textit{Doomism} and \textit{Local Experiential} show lower persistence in high-misinformation groups. For example, \textit{Doomism} decreases from 43.1 hours (Low) to 20.3 hours (High), while \textit{Local Experiential} increases from Low (30.5 hours) to Mid (37.3 hours) before decreasing at High (16.8 hours).


\begin{table}[t]
\centering
\small
\begin{tabular}{lrrr}
\hline
\textbf{Discourse Profile} &
\textbf{$H$ Statistic} &
\textbf{Raw $p$} &
\textbf{BH-FDR $p$} \\
\hline
Systemic Critique  & 20.501 & $<0.001$ & $\mathbf{0.003}$ \\
Local Experiential & 16.336 & $<0.001$ & $\mathbf{0.003}$ \\
Conspiracy         & 9.237  & 0.010    & $\mathbf{0.020}$ \\
Pro Science        & 4.875  & 0.047    & 0.071 \\
Political Combat   & 3.850  & 0.146    & 0.175 \\
Doomism            & 2.912  & 0.233    & 0.233 \\
\hline
\end{tabular}
\caption{Within-profile Kruskal--Wallis tests examining whether
persistence differs across misinformation tertiles (Low, Mid, High).
BH-FDR denotes Benjamini--Hochberg correction across the six
profile-level tests. Bold values remain significant at $\alpha=0.05$. After BH-FDR correction, persistence differs significantly across misinformation tertiles for Systemic Critique, Local Experiential, and Conspiracy.}
\vspace{-0.4em}
\label{tab:kruskal_wallis}
\end{table}

To assess whether these differences reflect statistically significant heterogeneity, we conduct within-profile Kruskal--Wallis tests comparing persistence across misinformation levels and apply Benjamini--Hochberg FDR correction across the six profile-level tests. Table~\ref{tab:kruskal_wallis} shows that, after correction, \textit{Systemic Critique} ($p_{\mathrm{adj}}=0.003$), \textit{Local Experiential} ($p_{\mathrm{adj}}=0.003$), and \textit{Conspiracy} ($p_{\mathrm{adj}}=0.020$) remain significant. \textit{Pro Science}, which is significant before correction ($p=0.047$), no longer reaches significance ($p_{\mathrm{adj}}=0.071$), while \textit{Political Combat} and \textit{Doomism} remain non-significant. Thus, after multiple-comparison correction, we find evidence of persistence differences across misinformation levels for three profiles, while such differences are not detected for the remaining three.

\section{Discussion and Conclusion}
This work demonstrates that discourse profile is a useful, interpretable axis for climate misinformation research. By modeling how comments are framed in addition to what is said, we show that discourse profiles capture discourse-level regularities that complement comment text.

Our classification experiments show the clearest benefits of discourse-profile conditioning for lightweight text representations. Under video-disjoint evaluation, the correct-profile models significantly improve macro-F1 across three training seeds for bag-of-words, TF-IDF, and frozen MPNet representations. The largest improvement occurs with TF-IDF, while the MPNet improvement is smaller. RoBERTa-large shows a positive average difference, but the effect varies substantially across seeds and does not reach statistical significance. We also show that discourse-profile-conditioned LLM agents can serve as controlled probes of misinformation susceptibility. Holding claims constant while varying discourse profile produces systematic differences in acceptance behavior that align with human discourse-profile-level vulnerability. This suggests that discourse profiles provide a meaningful abstraction not only for classification, but also for analyzing susceptibility patterns.

Finally, our temporal dynamics analysis shows that engagement persistence co-varies with misinformation prevalence differently across discourse profiles. Some discourse contexts exhibit higher persistence in higher-misinformation settings, while others show the opposite pattern. These heterogeneous associations suggest that misinformation prevalence and engagement persistence are not uniformly distributed across discourse contexts, highlighting the importance of accounting for rhetorical variation when analyzing climate misinformation dynamics.

Together, these findings suggest that discourse-profile-aware modeling offers a principled way to surface discourse-level structure in climate misinformation discourse, unifying prediction, robustness analysis, susceptibility probing, and engagement dynamics within a single abstraction. Rather than replacing text-centric approaches, discourse profiles complement them by capturing stable patterns of expression and interaction that are difficult to recover from isolated comments alone.

\clearpage
\section*{Limitations}
\paragraph{Discourse Profile Construction Validity and Subjectivity.}
Our discourse profiles are constructed via embedding-based k-means clustering followed by GPT-5-based consolidation of the 21 fine-grained clusters into six higher-level types. While this improves interpretability, the resulting taxonomy is not uniquely determined: different embedding models, clustering algorithms, random seeds, \(k\)-selection criteria, language models, or consolidation prompts could yield alternative groupings. In particular, the consolidation step is model- and prompt-dependent and should not be interpreted as producing a uniquely correct taxonomy. Accordingly, the resulting discourse profiles are best viewed as dataset-specific rhetorical groupings rather than definitive or exhaustive categories of users. The independent LLM reassignment analysis provides a supplemental reproducibility check, but it does not substitute for conventional human inter-annotator agreement.

We infer discourse profiles from observed comment text and do not have access to ground-truth user attributes (e.g., demographics, ideology, geography, or identity) or off-platform behavior. Consequently, discourse profile labels should not be interpreted as psychological traits or real-world subpopulations; they reflect recurring linguistic and rhetorical patterns expressed in public TikTok comments within our collection.

\paragraph{Cross-Domain Misinformation Label Transfer.}
Our misinformation labels on TikTok comments are produced by a classifier trained on \textsc{QuotaClimat} and transferred to in-the-wild \textsc{ClimateDisc} discourse. In addition to claim--comment matching and LLM-based agreement checks, we evaluate this transfer on 100 randomly sampled comments independently labeled by two human annotators (Appendix~\ref{app:human_validation}). Human--human agreement is substantial ($\kappa=0.76$), whereas classifier--human agreement is moderate ($\kappa=0.57$ and $0.55$). These results provide direct evidence about cross-domain label quality but also confirm non-negligible label noise. The \textsc{QuotaClimat} taxonomy is claim-centric and may not fully capture sarcasm, implicature, partial agreement, or conversational context in TikTok comments. We therefore treat individual transferred labels as imperfect and interpret downstream results primarily as aggregate profile-level associations.

\paragraph{Cross-Platform Generalization.}
The discourse profiles are constructed from a single TikTok climate-discourse corpus. We do not claim that the resulting six-profile taxonomy is platform invariant; rhetorical conventions and interaction patterns may differ across platforms, topics, and communities. Cross-platform profile construction and validation therefore remain important directions for future work.

\paragraph{LLM Agents Susceptibility.}

Our LLM-based susceptibility results should be interpreted as a controlled probe rather than a behavioral simulation of real users. Outputs can be sensitive to prompt wording, decoding settings, and the specific model used, and we evaluate only a single LLM configuration. Furthermore, our metric captures refusal/acceptance patterns under instruction rather than real-world exposure, incentives, or social context. We therefore emphasize alignment in relative discourse profile ordering rather than treating absolute LLM susceptibility rates as estimates of human vulnerability.

\paragraph{Temporal Dynamics and Causality.}
Our persistence analysis faces important methodological constraints. First, persistence hours conflate organic sharing velocity, algorithmic promotion, and content moderation decisions, precluding mechanistic attribution. Second, our cross-sectional design cannot establish whether misinformation density affects persistence or whether longer-persisting videos accumulate more misinformation over time. Third, video-level aggregation may obscure important dynamics at finer temporal scales.

\paragraph{Platform and Data Constraints.}
TikTok's proprietary recommendation algorithm and content moderation policies are opaque, limiting our ability to interpret persistence patterns mechanistically. We cannot observe deleted, shadowbanned, or algorithmically suppressed content, potentially biasing our sample toward platform-permitted content.

\paragraph{Ethical Considerations.}
Our analysis focuses on public comments and does not identify or track individual users. Nevertheless, discourse profile taxonomies could potentially be misinterpreted or misused as a form of user profiling. We emphasize that this is neither our intent nor what our method does. The discourse profiles in this work are comment-level abstractions constructed from recurring rhetorical patterns in public discourse, not demographic, psychological, or persistent individual traits. A user may express different discourse patterns across comments or contexts, and our analysis does not observe or infer real-world user identities, beliefs, or susceptibility. Accordingly, discourse profiles should be used to study aggregate discourse patterns and model behavior, not to target or make claims about individual users.

\section*{Acknowledgments}

\paragraph{Personally Identifiable / Offensive Information.}
We anonymize all potentially personally identifiable information in our released materials, including file paths and organization names in the code. In addition, examples of user comments shown in the paper have been sanitized to remove profane or offensive language.

\paragraph{AI Assistants in Writing.}
We used ChatGPT to assist with language polishing, \LaTeX{} formatting refinement, and code refactoring. Separately, language models were used as methodological or experimental components in GPT-5-based cluster consolidation, susceptibility probing, stylistic-rewrite generation, annotation, and supplemental profile-reassignment analyses described in the corresponding sections.

\bibliography{custom}

@inproceedings{gao-etal-2025-insights,
    title = "Insights into Climate Change Narratives: Emotional Alignment and Engagement Analysis on {T}ik{T}ok",
    author = "Gao, Ge  and
      Shan, Zhengyang  and
      Crissman, James  and
      Novozhilova, Ekaterina  and
      Huang, YuCheng  and
      Ramanathan, Arti  and
      Betke, Margrit  and
      Wijaya, Derry",
    editor = "Atwell, Katherine  and
      Biester, Laura  and
      Borah, Angana  and
      Dementieva, Daryna  and
      Ignat, Oana  and
      Kotonya, Neema  and
      Liu, Ziyi  and
      Wan, Ruyuan  and
      Wilson, Steven  and
      Zhao, Jieyu",
    booktitle = "Proceedings of the Fourth Workshop on NLP for Positive Impact (NLP4PI)",
    month = jul,
    year = "2025",
    address = "Vienna, Austria",
    publisher = "Association for Computational Linguistics",
    url = "https://aclanthology.org/2025.nlp4pi-1.11/",
    doi = "10.18653/v1/2025.nlp4pi-1.11",
    pages = "128--143",
    ISBN = "978-1-959429-19-7",
}

@ARTICLE{kmeans,
  author={Lloyd, S.},
  journal={IEEE Transactions on Information Theory}, 
  title={Least squares quantization in PCM}, 
  year={1982},
  volume={28},
  number={2},
  pages={129-137},
  doi={10.1109/TIT.1982.1056489}}

@misc{roberta,
      title={{RoBERTa}: A Robustly Optimized {BERT} Pretraining Approach}, 
      author={Yinhan Liu and Myle Ott and Naman Goyal and Jingfei Du and Mandar Joshi and Danqi Chen and Omer Levy and Mike Lewis and Luke Zettlemoyer and Veselin Stoyanov},
      year={2019},
      eprint={1907.11692},
      archivePrefix={arXiv},
      primaryClass={cs.CL},
      url={https://arxiv.org/abs/1907.11692}, 
}

@inproceedings{marsden2019,
    author = {Marsden, Nicola and Pröbster, Monika},
    year = {2019},
    month = {01},
    pages = {},
    title = {Personas and Identity Looking at Multiple Identities to Inform the Construction of Personas},
    booktitle = {Proceedings of the 2019 CHI Conference on Human Factors in Computing Systems},
    doi = {10.1145/3290605.3300565}
}

@article{article,
author = {Bloomfield, Emma Frances and Tillery, Denise},
year = {2018},
month = {12},
pages = {1-12},
title = {The Circulation of Climate Change Denial Online: Rhetorical and Networking Strategies on Facebook},
volume = {13},
journal = {Environmental Communication},
doi = {10.1080/17524032.2018.1527378}
}

@inproceedings{del-tredici-etal-2019-shall,
    title = "You Shall Know a User by the Company It Keeps: Dynamic Representations for Social Media Users in {NLP}",
    author = "Del Tredici, Marco  and
      Marcheggiani, Diego  and
      Schulte im Walde, Sabine  and
      Fern{\'a}ndez, Raquel",
    editor = "Inui, Kentaro  and
      Jiang, Jing  and
      Ng, Vincent  and
      Wan, Xiaojun",
    booktitle = "Proceedings of the 2019 Conference on Empirical Methods in Natural Language Processing and the 9th International Joint Conference on Natural Language Processing (EMNLP-IJCNLP)",
    month = nov,
    year = "2019",
    address = "Hong Kong, China",
    publisher = "Association for Computational Linguistics",
    url = "https://aclanthology.org/D19-1477/",
    doi = "10.18653/v1/D19-1477",
    pages = "4707--4717"
}

@inproceedings{barel-etal-2025-acquired,
    title = "Acquired {TASTE}: Multimodal Stance Detection with Textual and Structural Embeddings",
    author = "Barel, Guy  and
      Tsur, Oren  and
      Vilenchik, Dan",
    editor = "Rambow, Owen  and
      Wanner, Leo  and
      Apidianaki, Marianna  and
      Al-Khalifa, Hend  and
      Eugenio, Barbara Di  and
      Schockaert, Steven",
    booktitle = "Proceedings of the 31st International Conference on Computational Linguistics",
    month = jan,
    year = "2025",
    address = "Abu Dhabi, UAE",
    publisher = "Association for Computational Linguistics",
    url = "https://aclanthology.org/2025.coling-main.433/",
    pages = "6492--6504"
}

@inproceedings{lynn-etal-2019-tweet,
    title = "Tweet Classification without the Tweet: An Empirical Examination of User versus Document Attributes",
    author = "Lynn, Veronica  and
      Giorgi, Salvatore  and
      Balasubramanian, Niranjan  and
      Schwartz, H. Andrew",
    editor = "Volkova, Svitlana  and
      Jurgens, David  and
      Hovy, Dirk  and
      Bamman, David  and
      Tsur, Oren",
    booktitle = "Proceedings of the Third Workshop on Natural Language Processing and Computational Social Science",
    month = jun,
    year = "2019",
    address = "Minneapolis, Minnesota",
    publisher = "Association for Computational Linguistics",
    url = "https://aclanthology.org/W19-2103/",
    doi = "10.18653/v1/W19-2103",
    pages = "18--28"
}

@inproceedings{Qiu2015ModelingUA,
  title={Modeling User Arguments, Interactions, and Attributes for Stance Prediction in Online Debate Forums},
  author={Minghui Qiu and Yanchuan Sim and Noah A. Smith and Jing Jiang},
  booktitle={SDM},
  year={2015},
  url={https://api.semanticscholar.org/CorpusID:1011342}
}

@article{climateOnSocial1,
    author = {Pearce, Warren and Niederer, Sabine and Ozkula, Suay and Sanchez-Querubin, Natalia},
    year = {2018},
    month = {12},
    pages = {},
    title = {The social media life of climate change: Platforms, publics, and future imaginaries},
    volume = {10},
    journal = {Wiley Interdisciplinary Reviews: Climate Change},
    doi = {10.1002/wcc.569}
}

@article{climateOnSocial2,
    author = {Sultana, Bebe Chand and Prodhan, Md and Alam, Edris and Sohel, Md and Bari, A. B. M. Mainul and Pal, Subodh and Islam, Md and Islam, Abu},
    year = {2024},
    month = {09},
    pages = {1301400},
    title = {A systematic review of the nexus between climate change and social media: present status, trends, and future challenges},
    volume = {9},
    journal = {Frontiers in Communication},
    doi = {10.3389/fcomm.2024.1301400}
}

@article{climateOnTK1,
    author = {Basch, Corey and Yalamanchili, Bhavya and Fera, Joseph},
    title   = {\#Climate Change on TikTok: A Content Analysis of Videos},
    journal = {Journal of Community Health},
    year    = {2022},
    volume  = {47},
    number  = {1},
    pages   = {163--167},
    doi     = {10.1007/s10900-021-01031-x}
}

@inproceedings{climateOnTK2,
    author = {Pera, Arianna and Aiello, Luca},
    year = {2024},
    month = {05},
    pages = {376-381},
    title = {Shifting Climates: Climate Change Communication from YouTube to TikTok},
    booktitle = {Proceedings of the 16th ACM Web Science Conference},
    doi = {10.1145/3614419.3644024}
}

@inproceedings{tkMisinfo1,
    author       = {Clara Baltasar and Sergio D'Antonio Maceiras and Alejandro Martín and David Camacho},
    title        = {Analysis of Climate Change Misleading Information in TikTok},
    booktitle    = {Proceedings of the 1st Workshop on COuntering Disinformation with Artificial Intelligence (CODAI), co-located with the 27th European Conference on Artificial Intelligence (ECAI 2024)},
    series       = {CEUR Workshop Proceedings},
    volume       = {3782},
    pages        = {54--61},
    year         = {2024},
    publisher    = {CEUR-WS.org},
    address      = {Santiago de Compostela, Spain},
    note         = {Creative Commons CC BY 4.0},
    url          = {https://ceur-ws.org/Vol-3782/paper6short.pdf}
}

@article{priorClimate1,
    author = {Islam, Md Rafiqul and Liu, Shaowu and Wang, Xianzhi and Xu, Guandong},
    year = {2020},
    month = {09},
    pages = {},
    title = {Deep learning for misinformation detection on online social networks: a survey and new perspectives},
    volume = {10},
    journal = {Social Network Analysis and Mining},
    doi = {10.1007/s13278-020-00696-x}
}

@article{priorClimate2,
    author = {Coan, Travis and Boussalis, Constantine and Cook, John and Nanko, Mirjam Odile},
    year = {2021},
    month = {11},
    pages = {},
    title = {Computer-assisted classification of contrarian claims about climate change},
    volume = {11},
    journal = {Scientific Reports},
    doi = {10.1038/s41598-021-01714-4}
}

@article{priorClimate3,
    author = {Upadhyaya, Apoorva and Fisichella, Marco and Nejdl, Wolfgang},
    year = {2023},
    month = {07},
    pages = {103325},
    title = {Towards sentiment and Temporal Aided Stance Detection of climate change tweets},
    volume = {60},
    journal = {Information Processing \& Management},
    doi = {10.1016/j.ipm.2023.103325}
}

@misc{priorClimate4,
    doi = {10.21227/cdaz-jh77},
    url = {https://dx.doi.org/10.21227/cdaz-jh77},
    author = {YeonJung Choi and Lanyu Shang and Dong Wang},
    publisher = {IEEE Dataport},
    title = {ClimateMiSt: Climate Change Misinformation and Stance Detection Dataset},
    year = {2023} 
}

@article{priorClimate5,
    author = {Rojas, Cristian and Algra-Maschio, Frank and Andrejevic, Mark and Coan, Travis and Cook, John and Li, Yuan-Fang},
    year = {2024},
    month = {08},
    pages = {},
    title = {Hierarchical machine learning models can identify stimuli of climate change misinformation on social media},
    volume = {5},
    journal = {Communications Earth \& Environment},
    doi = {10.1038/s43247-024-01573-7}
}

@article{Song2020MPNetMA,
  title={MPNet: Masked and Permuted Pre-training for Language Understanding},
  author={Kaitao Song and Xu Tan and Tao Qin and Jianfeng Lu and Tie-Yan Liu},
  journal={ArXiv},
  year={2020},
  volume={abs/2004.09297},
  url={https://api.semanticscholar.org/CorpusID:215827489}
}

@article{Vosoughi2018,
author = {Soroush Vosoughi  and Deb Roy  and Sinan Aral },
title = {The spread of true and false news online},
journal = {Science},
volume = {359},
number = {6380},
pages = {1146-1151},
year = {2018},
doi = {10.1126/science.aap9559},
URL = {https://www.science.org/doi/abs/10.1126/science.aap9559},
eprint = {https://www.science.org/doi/pdf/10.1126/science.aap9559}}

@article{SUN2024108271,
title = {Who shares misinformation on social media? A meta-analysis of individual traits related to misinformation sharing},
journal = {Computers in Human Behavior},
volume = {158},
pages = {108271},
year = {2024},
issn = {0747-5632},
doi = {https://doi.org/10.1016/j.chb.2024.108271},
url = {https://www.sciencedirect.com/science/article/pii/S0747563224001390},
author = {Yanqing Sun and Juan Xie}
}

@article{Grinberg2019,
author = {Nir Grinberg  and Kenneth Joseph  and Lisa Friedland  and Briony Swire-Thompson  and David Lazer },
title = {Fake news on Twitter during the 2016 U.S. presidential election},
journal = {Science},
volume = {363},
number = {6425},
pages = {374-378},
year = {2019},
doi = {10.1126/science.aau2706},
URL = {https://www.science.org/doi/abs/10.1126/science.aau2706},
eprint = {https://www.science.org/doi/pdf/10.1126/science.aau2706}}

@inproceedings{benton-dredze-2018-using,
    title = "Using Author Embeddings to Improve Tweet Stance Classification",
    author = "Benton, Adrian  and
      Dredze, Mark",
    editor = "Xu, Wei  and
      Ritter, Alan  and
      Baldwin, Tim  and
      Rahimi, Afshin",
    booktitle = "Proceedings of the 2018 {EMNLP} Workshop W-{NUT}: The 4th Workshop on Noisy User-generated Text",
    month = nov,
    year = "2018",
    address = "Brussels, Belgium",
    publisher = "Association for Computational Linguistics",
    url = "https://aclanthology.org/W18-6124/",
    doi = "10.18653/v1/W18-6124",
    pages = "184--194"
}

@inproceedings{liu-etal-2023-analyzing,
    title = "Analyzing and Predicting Persistence of News Tweets",
    author = "Liu, Maggie  and
      Wang, Jing  and
      Preotiuc-Pietro, Daniel",
    editor = "Park, Jong C.  and
      Arase, Yuki  and
      Hu, Baotian  and
      Lu, Wei  and
      Wijaya, Derry  and
      Purwarianti, Ayu  and
      Krisnadhi, Adila Alfa",
    booktitle = "Proceedings of the 13th International Joint Conference on Natural Language Processing and the 3rd Conference of the Asia-Pacific Chapter of the Association for Computational Linguistics (Volume 1: Long Papers)",
    month = nov,
    year = "2023",
    address = "Nusa Dua, Bali",
    publisher = "Association for Computational Linguistics",
    url = "https://aclanthology.org/2023.ijcnlp-main.46/",
    doi = "10.18653/v1/2023.ijcnlp-main.46",
    pages = "719--732"
}

@article{lamb,
    author = {Lamb, William and Mattioli, Giulio and Levi, Sebastian and Roberts, J. and Capstick, Stuart and Creutzig, Felix and Minx, Jan and Müller-Hansen, Finn and Culhane, Trevor and Steinberger, Julia},
    year = {2020},
    month = {07},
    pages = {},
    title = {Discourses of climate delay},
    volume = {3},
    journal = {Global Sustainability},
    doi = {10.1017/sus.2020.13}
}

@article{holder,
    title={Climate obstruction and Facebook advertising: how a sample of climate obstruction organizations use social media to disseminate discourses of delay},
    author={Faye Holder and Sanober Mirza and Namson-Ngo-Lee and Jake Carbone and Ruth E. McKie},
    journal={Climatic Change},
    year={2023},
    volume={176},
    pages={1-21},
    url={https://api.semanticscholar.org/CorpusID:256701745}
}

@inproceedings{llm1,
    author = {Park, Joon and O'Brien, Joseph and Cai, Carrie and Morris, Meredith and Liang, Percy and Bernstein, Michael},
    year = {2023},
    month = {10},
    pages = {1-22},
    title = {Generative Agents: Interactive Simulacra of Human Behavior},
    booktitle = {Proceedings of the 36th Annual ACM Symposium on User Interface Software and Technology},
    doi = {10.1145/3586183.3606763}
}

@article{llm2,
    author = {Krishna, Ranjay and Lee, Donsuk and Fei-Fei, Li and Bernstein, Michael},
    year = {2022},
    month = {09},
    pages = {},
    title = {Socially situated artificial intelligence enables learning from human interaction},
    volume = {119},
    journal = {Proceedings of the National Academy of Sciences},
    doi = {10.1073/pnas.2115730119}
}

@article{llm3,
    author = {Argyle, Lisa and Busby, Ethan and Fulda, Nancy and Gubler, Joshua and Rytting, Christopher and Wingate, David},
    year = {2023},
    month = {02},
    pages = {1-15},
    title = {Out of One, Many: Using Language Models to Simulate Human Samples},
    volume = {31},
    journal = {Political Analysis},
    doi = {10.1017/pan.2023.2}
}

@inproceedings{llmpersona,
    title = "The steerability of large language models toward data-driven personas",
    author = "Li, Junyi  and
      Peris, Charith  and
      Mehrabi, Ninareh  and
      Goyal, Palash  and
      Chang, Kai-Wei  and
      Galstyan, Aram  and
      Zemel, Richard  and
      Gupta, Rahul",
    editor = "Duh, Kevin  and
      Gomez, Helena  and
      Bethard, Steven",
    booktitle = "Proceedings of the 2024 Conference of the North American Chapter of the Association for Computational Linguistics: Human Language Technologies (Volume 1: Long Papers)",
    month = jun,
    year = "2024",
    address = "Mexico City, Mexico",
    publisher = "Association for Computational Linguistics",
    url = "https://aclanthology.org/2024.naacl-long.405/",
    doi = "10.18653/v1/2024.naacl-long.405",
    pages = "7290--7305",
}

@misc{QuotaClimat,
  author = {QuotaClimat},
  title = {Frugal AI Challenge Text Train Dataset},
  year = {2025},
  publisher = {Hugging Face},
  url = {https://huggingface.co/datasets/QuotaClimat/frugalaichallenge-text-train},
  note = {Dataset for the Frugal AI Challenge 2025, focused on climate disinformation narratives}
}

@misc{all-mpnet-base-v2,
  author = {Reimers, Nils and Gurevych, Iryna},
  title = {all-mpnet-base-v2: Sentence-Transformers Model},
  year = {2020},
  publisher = {Hugging Face},
  url = {https://huggingface.co/sentence-transformers/all-mpnet-base-v2},
  note = {Fine-tuned for sentence embeddings on 1B+ sentence pairs}
}

@inproceedings{nguyen2020fang,
    title={FANG: Leveraging Social Context for Fake News Detection Using Graph Representation},
   url={http://dx.doi.org/10.1145/3340531.3412046},
   DOI={10.1145/3340531.3412046},
   booktitle={Proceedings of the 29th ACM International Conference on Information \&; Knowledge Management},
   publisher={ACM},
   author={Nguyen, Van-Hoang and Sugiyama, Kazunari and Nakov, Preslav and Kan, Min-Yen},
   year={2020},
   month=Oct, pages={1165–1174},
   collection={CIKM ’20} }

@inproceedings{mehta2022gnn,
  title = "Tackling Fake News Detection by Continually Improving Social Context Representations using Graph Neural Networks",
    author = "Mehta, Nikhil  and
      Pacheco, Maria Leonor  and
      Goldwasser, Dan",
    editor = "Muresan, Smaranda  and
      Nakov, Preslav  and
      Villavicencio, Aline",
    booktitle = "Proceedings of the 60th Annual Meeting of the Association for Computational Linguistics (Volume 1: Long Papers)",
    month = may,
    year = "2022",
    address = "Dublin, Ireland",
    publisher = "Association for Computational Linguistics",
    url = "https://aclanthology.org/2022.acl-long.97/",
    doi = "10.18653/v1/2022.acl-long.97",
    pages = "1363--1380"
}

@inproceedings{allaway2021zero,
  title={Zero-Shot Stance Detection: A Dataset and Model Analysis},
  author={Allaway, Emily and McKeown, Kathleen},
  booktitle={NAACL},
  year={2021}
}

@misc{zhang2023doublehtwitteruserstance,
      title={DoubleH: Twitter User Stance Detection via Bipartite Graph Neural Networks}, 
      author={Chong Zhang and Zhenkun Zhou and Xingyu Peng and Ke Xu},
      year={2023},
      eprint={2301.08774},
      archivePrefix={arXiv},
      primaryClass={cs.SI},
      url={https://arxiv.org/abs/2301.08774}, 
}

@article{tfidf,
    author = {Jones, K.},
    year = {2004},
    month = {01},
    pages = {493-502},
    title = {A Statistical Interpretation of Term Specificity in Retrieval},
    volume = {60},
    journal = {Journal of Documentation},
    doi = {10.1108/00220410410560573}
}

@inproceedings{mpcg,
    title = "{MPCG}: Multi-Round Persona-Conditioned Generation for Modeling the Evolution of Misinformation with {LLM}s",
    author = "Brian, Chong Jun Rong  and
      Tang, Yixuan  and
      Tung, Anthony Kum Hoe",
    editor = "Christodoulopoulos, Christos  and
      Chakraborty, Tanmoy  and
      Rose, Carolyn  and
      Peng, Violet",
    booktitle = "Proceedings of the 2025 Conference on Empirical Methods in Natural Language Processing",
    month = nov,
    year = "2025",
    address = "Suzhou, China",
    publisher = "Association for Computational Linguistics",
    url = "https://aclanthology.org/2025.emnlp-main.1727/",
    doi = "10.18653/v1/2025.emnlp-main.1727",
    pages = "34030--34064",
    ISBN = "979-8-89176-332-6",
}

\clearpage
\appendix

\section{Taxonomy Transfer Details}
\label{appendix:taxonomy_transfer}
This appendix describes how we obtain fine-grained misinformation-type labels for in-the-wild comments by transferring a curated taxonomy to the \textsc{ClimateDisc} corpus.

\subsection{Misinformation Classifier Training}
We fine-tune a RoBERTa-large classifier on \textsc{QuotaClimat} to predict one of eight climate misinformation categories from text. Training is performed on a single NVIDIA A100 GPU for four epochs with batch size 128 and learning rate \(2\times10^{-5}\). Our model reaches a test accuracy of 0.75 on the eight-class \textsc{QuotaClimat} classification task. This provides a source-domain performance reference; because the downstream application is to in-the-wild TikTok comments, we separately evaluate cross-domain agreement below.

\subsection{Cross-domain Validation via Claim-Comment Matching}
To validate transfer from curated claims to in-the-wild comments, we embed both \textsc{QuotaClimat} claims and \textsc{ClimateDisc} comments using the same sentence transformer and identify high-similarity claim--comment pairs via cosine similarity. Pairs exceeding a fixed threshold of $\tau=0.85$ are treated as referring to the same underlying claim. We select $\tau=0.85$ via qualitative inspection: it is the lowest value that consistently yields pairs about the same underlying claim with aligned stance; lower thresholds increasingly admit topic-level matches with stance mismatches. This produces 208 matched comment--claim pairs (Table~\ref{tab:matched_pairs}).

On these pairs, the classifier's prediction on the TikTok comment matches the ground-truth label of the paired \textsc{QuotaClimat} claim 79\% of the time (majority baseline 38\%), with Cohen's $\kappa=0.72$ indicating strong agreement beyond chance (Table \ref{tab:kappa_interpret}).
Manual inspection and quantitative agreement metrics further support that matched pairs refer to the same underlying claim. 

\begin{table}[t]
\centering
\small
\begin{tabular}{|p{0.45\linewidth}|p{0.45\linewidth}|}
\hline
\textbf{Comment (ClimateDisc)} & \textbf{Matched Claim (QuotaClimat)} \\
\hline

Climate change is a HOAX
&
Climate change is a complete hoax always has been.
\\
\hline

Yes, climate change is real to what extent is arguable, but definitely changing
&
Climate change is definitely real. It changes every year from winter to autumn to spring to summer.
\\
\hline

Climate change is very real. The climate has always changed, but in recent years it is changing too fast
&
Climate change is definitely real. It changes every year from winter to autumn to spring to summer.
\\
\hline
\end{tabular}
\caption{Representative matched comment--claim pairs obtained using a cosine-similarity threshold of 0.85.}
\label{tab:matched_pairs}
\end{table}

\begin{table}[t]
\centering
\begin{tabular}{|l|c|}
\hline
\textbf{Cohen's Kappa} & \textbf{Quality} \\ \hline
\texttt{>0.8}  & Almost Perfect Agreement        \\ \hline
\texttt{>0.6}  & Substantial Agreement         \\ \hline
\texttt{>0.4}  & Moderate Agreement         \\ \hline
\texttt{>0.2}  & Fair Agreement         \\ \hline
\texttt{0-0.2} & Slight Agreement         \\ \hline
\texttt{<0}    & Almost No Agreement         \\ \hline
\end{tabular}
\caption{Interpretation for different ranges of the Cohen's Kappa values.}
\label{tab:kappa_interpret}
\end{table}

Based on this validation, we apply the classifier to all comments in \textsc{ClimateDisc} to characterize misinformation types in in-the-wild discourse.

To assess label quality beyond the high-similarity regime, where explicit claim alignment is unavailable, we conduct an additional agreement analysis using an independent LLM annotator. We randomly sample 200 comments from \textsc{ClimateDisc} that do not exceed the similarity threshold with any \textsc{QuotaClimat} claim. Each comment is annotated independently by ChatGPT using a fixed prompt (Appendix~\ref{app:misinfo_agreement}), and we compare these annotations with our classifier’s predictions.

Across these unmatched comments, agreement reaches 89\%, with Cohen's $\kappa=0.54$. The high raw agreement but moderate $\kappa$ provides an additional cross-domain agreement check while also indicating that substantial uncertainty remains. Because an LLM comparison does not substitute for human annotation, we complement this analysis with an independent human validation study below.

\subsection{Cross-Domain Human Validation}
\label{app:human_validation}

\begin{table}[t]
\centering
\small
\begin{tabular}{lc}
\toprule
\textbf{Comparison} & \textbf{Cohen's $\kappa$} \\
\midrule
Human 1 $\leftrightarrow$ Human 2 & 0.76 \\
Human 1 $\leftrightarrow$ classifier & 0.57 \\
Human 2 $\leftrightarrow$ classifier & 0.55 \\
\bottomrule
\end{tabular}
\caption{Cross-domain agreement on 100 randomly sampled
\textsc{ClimateDisc} comments independently labeled by two human
annotators.}
\label{tab:human_validation}
\end{table}

To directly assess label quality on in-the-wild TikTok discourse, we randomly sample 100 comments from \textsc{ClimateDisc}. Two human annotators independently assign each comment one of the eight \textsc{QuotaClimat} categories using the same category definitions as the classifier. We assess both human--human agreement and agreement between each human annotator and the transferred classifier predictions using Cohen's $\kappa$.

The two human annotators show substantial agreement ($\kappa=0.76$), indicating that the taxonomy can be applied with reasonable consistency to the sampled in-the-wild TikTok comments. Agreement between the classifier and each human annotator is moderate ($\kappa=0.57$ and $0.55$). These results provide direct human evidence about cross-domain label quality while also confirming that non-negligible label noise remains. We therefore treat the \textsc{ClimateDisc} misinformation labels as automatically transferred labels rather than gold-standard comment-level annotations.

\section{Datasets}
\subsection{Climate Misinformation Types}
Below are the types of Climate Change Misinformation we used in \textsc{QuotaClimat}.
\label{appendix:disinfo}

\begin{description}
    \item[0\_not\_relevant:] No relevant claim detected or claims that don't fit other categories.
    \item[1\_not\_happening:] Claims denying the occurrence of global warming and its effects - Global warming is not happening. Climate change is NOT leading to melting ice (such as glaciers, sea ice, and permafrost), increased extreme weather, or rising sea levels. Cold weather also shows that climate change is not happening.
    \item[2\_not\_human:] Claims denying human responsibility in climate change - Greenhouse gases from humans are not the causing climate change.
    \item[3\_not\_bad:] Claims minimizing or denying negative impacts of climate change - The impacts of climate change will not be bad and might even be beneficial.
    \item[4\_solutions\_harmful\_unnecessary:] Claims against climate solutions - Climate solutions are harmful or unnecessary.
    \item[5\_science\_is\_unreliable:] Claims questioning climate science validity - Climate science is uncertain, unsound, unreliable, or biased.
    \item[6\_proponents\_biased:] Claims attacking climate scientists and activists - Climate scientists and proponents of climate action are alarmist, biased, wrong, hypocritical, corrupt, and/or politically motivated.
    \item[7\_fossil\_fuels\_needed:] Claims promoting fossil fuel necessity - We need fossil fuels for economic growth, prosperity, and to maintain our standard of living.
\end{description}

\subsection{Representative Comments of Each Cluster}
\label{appendix:samples}
Below are the top ten comments closest to each cluster centroid.

\normalsize
\textbf{Cluster 0} (Partisan Anti-Trump/Anti-GOP Rants):
\tiny
\begin{enumerate}\itemsep0pt \parskip0pt \parsep0pt
\item He’s a complete delusional fool, druggy, grapist conspiracy theorist, worm brain lunatic. This POS will never be sworn in. His whole family disowned this crazed weed [clown emoji].
\item lol u talkin bout Biden?
\item this guy is full of s*** Trump's been helping out more for the American people then Joe Biden and Kamala
\item My god Vance just can’t answer without lying, there is no way that man should be anywhere near the White House, he’s pathetic [skull emoji].
\item He’s clueless and all his facts. He’s the most uneducated person we ever had.
\item This man is really ignorant. He doesn’t understand what he’s even talking about.
\item Again, WTF?! He is such a moron!!
\item He is part of the problem and can’t understand his days are over and his grandchildren will suffer for his actions.
\item Trump is so stupid
\item He is so vile and evil!! Typical conman!
\end{enumerate}

\normalsize
\textbf{Cluster 1} (Local Weather Anecdotes):
\tiny
\begin{enumerate}\itemsep0pt \parskip0pt \parsep0pt
\item you know you right Cuz. it's unseasonably warm where I live too [laughing emoji].
\item This winter in Michigan there was only snow for a few weeks in January and now it's 50 degrees and sunny.
\item It was concerning in Michigan yesterday (60s) then it snowed today [laughing emoji].
\item I'm in Wisconsin and it was like 90 last week and now it's in the 70s/50s.
\item We’re having an unseasonably warm winter here in Minnesota.
\item we went from rainy to like 65 degrees, what is this summer?
\item It’s actually scary the climate is warm right now, especially since I am Canadian and not used to this [sweat smile emoji].
\item In Utah we hit record lows in November, I’m so over it [crying emoji].
\item it’s nice right now but Thursday/Friday will be almost 80 in Massachusetts [crying emoji] damn you global warming [victory hand emoji].
\item south west area is getting record rain and snow.
\end{enumerate}

\normalsize
\textbf{Cluster 2} (Emoji/Applause Spam — converted):

\tiny
\raggedright
\newcommand{\emo}[1]{[#1]\allowbreak}

\begin{enumerate}\itemsep0pt \parskip0pt \parsep0pt
\item TRUMP \emo{100 emoji}\emo{100 emoji}\emo{100 emoji}\emo{100 emoji}\emo{100 emoji}
\item \emo{blue heart emoji}\emo{blue heart emoji}\emo{blue heart emoji}\emo{blue heart emoji} P
\item sindio \emo{laughing emoji}\emo{laughing emoji}\emo{laughing emoji}
\item \emo{thumbs up emoji}\emo{thumbs up emoji}\emo{ok emoji}\emo{party emoji}
\item \emo{target emoji}\emo{target emoji}\emo{fire emoji}\emo{fire emoji}\emo{fire emoji}
\item \emo{100 emoji}\emo{100 emoji}\emo{100 emoji}\emo{blue heart emoji}\emo{blue heart emoji}
\item \emo{screaming emoji} unvorstellbar
\item \emo{thumbs up emoji}\emo{thumbs up emoji}\emo{clap emoji x many}
\item \emo{clap emoji}\emo{clap emoji}\emo{clap emoji}\emo{clap emoji}
\item \emo{facepalm emoji}\emo{facepalm emoji}\emo{facepalm emoji}
\end{enumerate}

\normalsize
\textbf{Cluster 3} (Minimal Agreement Tokens):
\tiny
\begin{enumerate}\itemsep0pt \parskip0pt \parsep0pt
\item Ya
\item ya
\item Right
\item Right
\item RIGHT
\item Right
\item Right
\item Right
\item Right
\item RIGHT
\end{enumerate}

\normalsize
\textbf{Cluster 4} (Consumption \& Plastics / Recycling):
\tiny
\begin{enumerate}\itemsep0pt \parskip0pt \parsep0pt
\item clothes or items that can’t be recycled get buried or dumped.
\item Stop buying plastic whenever possible.
\item I wish I could buy products not completely covered in plastic.
\item reuse, repurpose, buy less, switch to glass or metal; 2024 no new buying challenge.
\item so much usable stuff goes to the dump.
\item buying second hand clothes to save things from landfill [sweat smile emoji].
\item buying reusable plastic to replace paper is not a good idea.
\item milk in plastic jugs, bread in plastic bags, lunch meat in plastic [laughing emoji].
\item all ends up in landfill anyway.
\item I'm sure they don't even recycle any of it.
\end{enumerate}

\normalsize
\textbf{Cluster 5} (Polite Gratitude Replies):
\tiny
\begin{enumerate}\itemsep0pt \parskip0pt \parsep0pt
\item Thank you!
\item Thank you!
\item Thank you!
\item Thank you!
\item Thank you!
\item Thank you!
\item Thank you!
\item Thank you!
\item thank you!
\item Thank you!
\end{enumerate}

\normalsize
\textbf{Cluster 6} (Short Emotional Reactions):
\tiny
\begin{enumerate}\itemsep0pt \parskip0pt \parsep0pt
\item So true [laughing emoji]
\item So cool
\item So scary [crying face emoji]
\item SO TRUE [eyes emoji]
\item So happy [heart emoji]
\item So sad
\item So cute
\item So sad [broken heart emoji]
\item so nice
\item Boosting [blue heart emoji]
\end{enumerate}

\normalsize
\textbf{Cluster 7} (Explainers + Conspiracies):
\tiny
\begin{enumerate}\itemsep0pt \parskip0pt \parsep0pt
\item It’s science happens every 10 years and it used to be all water.
\item thermal expansion and added water from land ice.
\item Not a man made phenomenon.
\item land ice and thermal expansion.
\item nothing to do with spraying of skies and geoengineering.
\item water freezes and expands in limestone zones.
\item before this it was acid rain.
\item local event caused by warming Arctic waters.
\item reduced emissions reduced reflective particles, warming the ocean.
\item happens when exiting an ice age.
\end{enumerate}

\normalsize
\textbf{Cluster 8} (Doomism / Earth Fine, Humans Won’t):
\tiny
\begin{enumerate}\itemsep0pt \parskip0pt \parsep0pt
\item another great flood, ice age, warming cycle coming.
\item ecological and societal collapse is on the way.
\item the earth will rid itself of us.
\item earth will adapt; we will not survive.
\item Science people, we are doomed.
\item near the environmental point of no return.
\item I’m worried about mankind’s future.
\item we will die before it affects earth.
\item perilous times like never before.
\item we must wake up and restore habitability.
\end{enumerate}

\normalsize
\textbf{Cluster 9} (Excited Agreement Bursts):
\tiny
\begin{enumerate}\itemsep0pt \parskip0pt \parsep0pt
\item Omg yes
\item Yessss oh my god
\item yesssss
\item OMG
\item You said it
\item Yesss omg
\item OMG
\item Nailed it
\item Yes Eva
\item yes love that
\end{enumerate}

\normalsize
\textbf{Cluster 10} (Shared Anxiety \& Coping):
\tiny
\begin{enumerate}\itemsep0pt \parskip0pt \parsep0pt
\item I’m so sad and afraid [blue heart emoji][strong emoji].
\item we must be aware but also able to live; maybe you’re here to bring change.
\item Say a prayer [pray emoji].
\item I can't go on like normal.
\item always have hope.
\item I’m coping too.
\item same, persist with compassion.
\item same, I feel terrible.
\item I'm scared and sad, I don't want to live in trying times.
\item Me too, it’s too much.
\end{enumerate}

\normalsize
\textbf{Cluster 11} (Government Blame / Anarchy Talk):
\tiny
\begin{enumerate}\itemsep0pt \parskip0pt \parsep0pt
\item are you serious? it's our governments.
\item idiots in charge do nothing.
\item kicking indigenous peoples off lands.
\item governments don’t see this as a problem.
\item manipulation of the weak.
\item HAARP government control.
\item this is what anarchy is, people helping each other.
\item fight through organizing and political action.
\item tell political leaders, not us.
\item US government and capitalism accelerated this.
\end{enumerate}

\normalsize
\textbf{Cluster 12} (Storms \& Flooding):
\tiny
\begin{enumerate}\itemsep0pt \parskip0pt \parsep0pt
\item back-to-back storms caused flooding.
\item climatologists warned for decades.
\item storms stronger and more frequent.
\item flash flooding moving north.
\item happened in less than an hour.
\item regular rain causing floods.
\item houses awaiting repairs from hurricanes.
\item NYC flooding commentary.
\item unprecedented Midwest flooding.
\item every US region has natural disasters.
\end{enumerate}

\normalsize
\textbf{Cluster 13} (Geo-Self Mentions):
\tiny
\begin{enumerate}\itemsep0pt \parskip0pt \parsep0pt
\item I’m in Tennessee.
\item you're in my state now.
\item Pennsylvania, same dread.
\item be glad you're not in FL.
\item I live in Visalia.
\item don’t say this, I live there.
\item I live here, first time hearing this.
\item Southern Illinois.
\item here in PA too.
\item I'm in Oregon.
\end{enumerate}

\normalsize
\textbf{Cluster 14} (Pro-Science / Anti-Misinformation):
\tiny
\begin{enumerate}\itemsep0pt \parskip0pt \parsep0pt
\item proves you don’t know what you’re talking about.
\item you say peer reviewed studies are wrong.
\item misinformation spreads easily.
\item not okay to think yourself smarter than scientists.
\item will never do their own research.
\item you trust a convicted felon serial grapist? got it.
\item you have no idea but skeptics are dumb?
\item simple rebuke is always wrong.
\item these people don’t believe in science.
\item knowledge kills their arguments.
\end{enumerate}

\normalsize
\textbf{Cluster 15} (@-Mentions, anonymized):
\tiny
\begin{enumerate}\itemsep0pt \parskip0pt \parsep0pt
\item @s********
\item @p************
\item @R*******
\item @e******
\item @M*******
\item @H******
\item @e******
\item @R***********
\item @a*****
\item @x************
\end{enumerate}

\normalsize
\textbf{Cluster 16} (Climate Denial / Manipulation):
\tiny
\begin{enumerate}\itemsep0pt \parskip0pt \parsep0pt
\item Climate change is a joke. You are brainwashed.
\item THERE IS NO SUCH THING AS CLIMATE CHANGE!!
\item f u and climate change it's a f**k** farce; you must be an a**h*** to believe it.
\item Cut the crap. Nobody believes that mess.
\item Climate change is a make-believe crisis.
\item No such thing as climate change.
\item It's weather manipulation.
\item climate change is a hoax and scam.
\item finally someone tells the truth about climate.
\item our idea of climate change is a lie.
\end{enumerate}

\normalsize
\textbf{Cluster 17} (Energy \& Policy Debate):
\tiny
\begin{enumerate}\itemsep0pt \parskip0pt \parsep0pt
\item energy corporations greenwashing rhetoric.
\item ignoring massive fossil fuel emissions.
\item electric car carbon credits rant.
\item new refineries would be more efficient.
\item switch to renewables and nuclear.
\item promoting “clean” coal and gas.
\item outsourcing pollution abroad.
\item no more fossil fuel projects.
\item build nuclear if you care about environment.
\item greener energy benefits future generations.
\end{enumerate}

\normalsize
\textbf{Cluster 18} (Electoral Mobilization):
\tiny
\begin{enumerate}\itemsep0pt \parskip0pt \parsep0pt
\item Republicans are delusional, vote blue.
\item Vote straight blue ticket.
\item Vote blue up and down the ballot.
\item must vote Harris 2024.
\item Vote Blue.
\item Republican voting blue across the board.
\item Vote blue, this is sickening.
\item Vote red, blue is crazy.
\item Vote all Democrats.
\item Vote blue across the board.
\end{enumerate}

\normalsize
\textbf{Cluster 19} (Food / Agriculture \& Diet):
\tiny
\begin{enumerate}\itemsep0pt \parskip0pt \parsep0pt
\item need organic and regenerative farming.
\item rainforest deforestation due to animal agriculture.
\item Farming must stop but I need plant-based foods.
\item raising fuel instead of food.
\item plant-based diet is bs.
\item want to work in sustainable agriculture.
\item kill animals to grow plants.
\item forests cleared for soybean fields.
\item our habits destroy rainforests.
\item plant trees.
\end{enumerate}

\normalsize
\textbf{Cluster 20} (Anti-Corporation / Anti-Capitalism):
\tiny
\begin{enumerate}\itemsep0pt \parskip0pt \parsep0pt
\item we cannot count on corporations and the wealthy.
\item fundamental problem with capitalism.
\item corporations keep you poor.
\item profit motive stands in the way.
\item corporations get tax breaks.
\item wealthy keep getting richer.
\item someone is buying what corporations sell.
\item greedflation from corporations.
\item capitalism is a plague.
\item corporations do not care about the people.
\end{enumerate}

\section{Cluster and Discourse-Profile Statistics}
\label{app:cluster_dist}
\normalsize

\subsection{Cluster Validation}
\label{app:cluster_validation}

Intrinsic validation provides complementary evidence for this representation. The 14 retained clusters have mean lexical coherence of NPMI = 0.13, compared with -0.03 for the seven Noise / Low-Signal clusters. 

Across ten random initializations of $k$-means at $k=21$, mean pairwise ARI and NMI are 0.741 and 0.815, respectively, and average profile assignment agreement after consolidation is 82.8\%.

\paragraph{Independent profile reassignment.}
As a supplemental reproducibility check, Claude Opus 4.8 and GPT 5.5 independently reassigned the 21 clusters to the six discourse profiles or Noise / Low-Signal, without access to the original mapping in Table~\ref{tab:personas}. The two raters agree on 19 of 21 cluster assignments ($\kappa=0.88$). GPT agrees with the consolidation on 16/21 clusters ($\kappa=0.71$), while Claude agrees on 15/21 ($\kappa=0.64$). Because these additional raters are LLMs rather than human annotators, we treat this as a supplemental reproducibility check rather than conventional human inter-annotator agreement.

\begin{figure}[t]
\centering
\includegraphics[width=1\columnwidth]{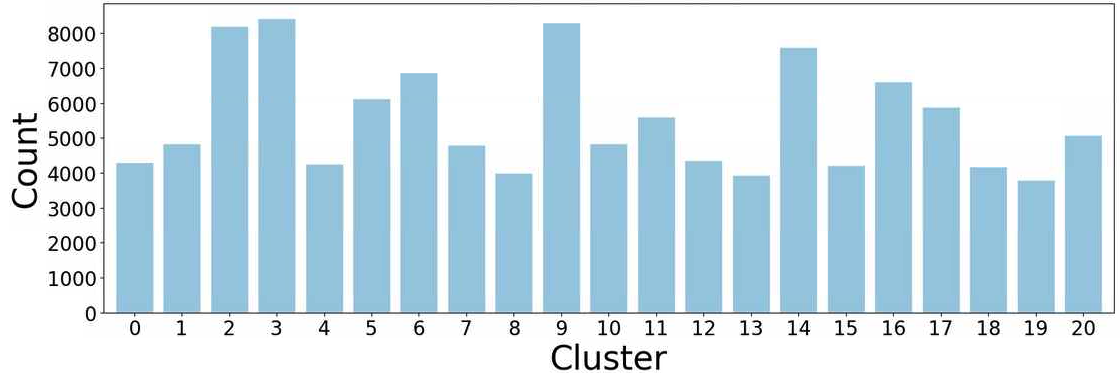} 
\includegraphics[width=1\columnwidth]{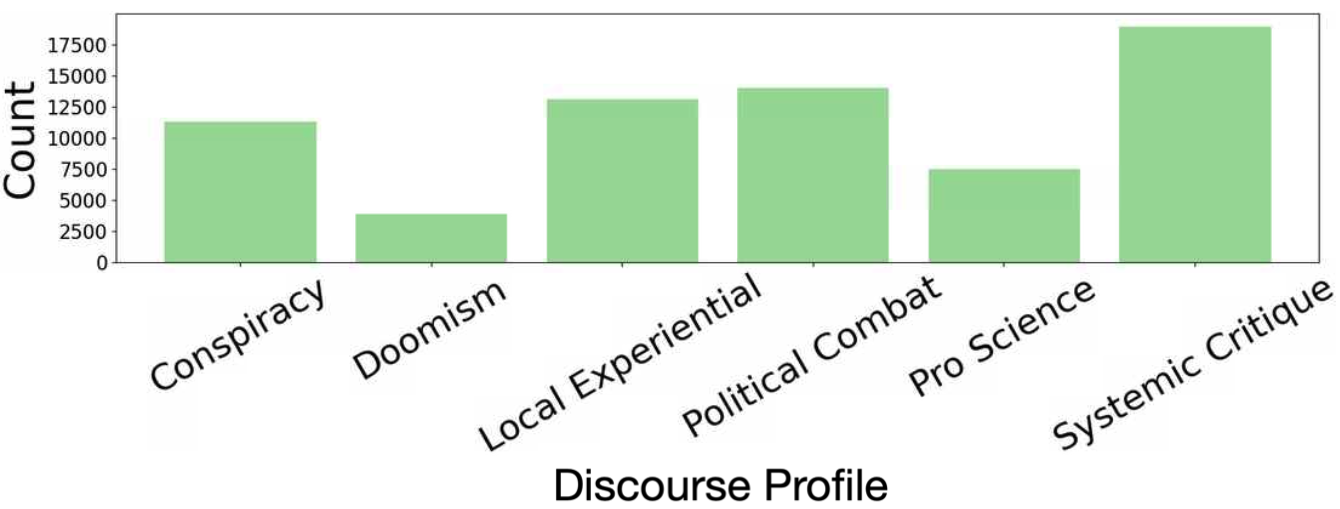} 
\caption{Distribution of cluster/discourse profile sizes.}
\label{clusterdist}
\vspace{-0.4em}
\end{figure}

Figure~\ref{clusterdist} shows that cluster and discourse profile sizes are relatively balanced: clusters range from 3{,}798--8{,}426 comments and discourse profiles from 3{,}992--19{,}019 comments. This reduces the risk that downstream effects are driven by a single dominant group.

\subsection{Comparison with HDBSCAN}
\label{app:hdbscan}

We implemented HDBSCAN on the same embeddings. HDBSCAN produced only six clusters while assigning 90.2\% of comments to noise, leaving only 9.8\% of the corpus available for downstream profile construction. Moreover, the retained comments collapsed almost entirely into two discourse profiles rather than recovering the broader taxonomy obtained with k-means. 

HDBSCAN is designed for settings where leaving a substantial portion of observations unassigned as noise is desirable. Our objective is different: we seek an exhaustive partition of the corpus into fine-grained semantic clusters that can subsequently be consolidated into broader discourse profiles. Accordingly, we retain k-means, which provides complete corpus coverage while producing interpretable fine-grained clusters that can be consolidated into reusable discourse profiles.

\section{Category-Level Classification Error Analysis}
\label{app:error_analysis}
The category-level analysis in Table \ref{tab:category_delta_f1} shows that the aggregate gains are not driven by a single misinformation label. For the lexical models, the largest improvements occur on categories that are comparatively difficult for the text-only baseline. Bag-of-words improves most on Science is Unreliable (+6.8 F1 points) and Not Bad (+5.7), while TF-IDF shows particularly large gains on Science is Unreliable (+23.4), Not Bad (+11.6), and Fossil Fuels Needed (+8.6). MPNet exhibits smaller but broadly positive changes, with its largest improvement on Fossil Fuels Needed (+3.6). RoBERTa-large likewise improves six of seven categories on average, most notably Science is Unreliable (+5.4), although its overall profile-aware gain is not statistically significant across training seeds.

\begin{table}[t]
\centering
\small
\setlength{\tabcolsep}{3pt}
\begin{tabular}{@{}lcccc@{}}
\toprule
\textbf{Category} & \textbf{BOW} & \textbf{TF-IDF} & \textbf{MPNet} & \textbf{RoBERTa} \\
\midrule
Science Unreliable
& +6.8 & +23.4 & +0.5 & +5.4 \\
Not Bad
& +5.7 & +11.6 & +1.8 & +0.8 \\
Not Happening
& +4.3 & +3.2 & +0.6 & +0.8 \\
Not Human-Caused
& +2.5 & +2.6 & +1.4 & $-$0.2 \\
Fossil Fuels Needed
& +1.2 & +8.6 & +3.6 & +1.3 \\
Solutions Harmful
& +0.7 & +4.0 & +0.0 & +1.7 \\
Proponents Biased
& $-$0.2 & +2.5 & +1.5 & +1.0 \\
\bottomrule
\end{tabular}
\caption{Mean category-level F1 change from the text-only model to the correct-profile model, averaged across three training seeds. The misinformation categories in this analysis are inherited from the \textsc{QuotaClimat} taxonomy. Values are F1 percentage-point differences. RoBERTa denotes RoBERTa-large.}
\label{tab:category_delta_f1}
\end{table}

\section{LLM-generated Stylistic Variants}
\label{app:llm_variants_section}

To evaluate robustness under stylistic variation, we additionally construct LLM-generated rewrites of misinformation comments following the multi-round rewriting setup of \citet{mpcg}. Starting from an original comment, we generate three successive rounds of rewrites using Llama-3.1-8B-Instruct. Each rewrite is conditioned on (i) a target discourse profile description, (ii) the original misinformation label definition, (iii) the associated video transcript context, and (iv) previously generated variants. The model is instructed to preserve the underlying misinformation category while varying rhetorical style and surface realization according to the target discourse profile.

For each original comment, variants are generated separately for each target discourse profile. The resulting Round-3 rewrites therefore represent progressively transformed stylistic variants of the original claim while retaining its misinformation category.

We evaluate both the text-only RoBERTa-large baseline (Model~A) and the discourse-profile-aware RoBERTa-large late-fusion model (Model~B) on these rewritten comments. Performance decreases for both models as rewriting progresses, consistent with prior findings that iterative LLM rewriting can degrade misinformation classification performance \cite{mpcg}. On the Round-3 rewrites, Model~A achieves 0.527 accuracy and 0.385 macro-F1, whereas Model~B achieves 0.638 accuracy and 0.534 macro-F1. Incorporating discourse profile therefore improves accuracy by 0.110 and macro-F1 by 0.149 under the strongest rewriting condition.

These results suggest that discourse-level behavioral patterns remain more stable than surface lexical and stylistic cues, allowing discourse-profile-aware models to better preserve predictive performance even when misinformation claims undergo substantial stylistic rewriting.

\section{LLM Prompts}
\subsection{Cluster Consolidation}
\label{app:cluster_consolidation}

To consolidate the 21 fine-grained semantic clusters into higher-level discourse profiles, we provided GPT-5 with the following instruction, followed by representative comments nearest each cluster centroid:

\begin{quote}
\small
\ttfamily
\raggedright

You are given 21 fine-grained semantic clusters derived from climate-change comments, each represented by comments nearest its cluster centroid. Group these clusters into a smaller set of higher-level discourse profiles according to shared rhetorical patterns and communicative stances toward climate change. For each higher-level discourse profile, provide a concise profile label and identify which of the 21 clusters belong to it.

\end{quote}

\subsection{Discourse Profile Card Construction}
\label{app:persona_card_construction}

\normalsize
To construct discourse profile cards, we prompt ChatGPT to synthesize coherent, reusable discourse profile descriptions from clusters of real-world comments. Each cluster is represented by a set of example comments (in Appendix \ref{appendix:samples}) drawn from our dataset, along with a short cluster label summarizing its dominant theme.

The following prompt template is used to generate each discourse profile card. The cluster label and representative comments are injected into the \texttt{\{cluster\_description\}} field.

\begin{quote}
\small
\ttfamily
\raggedright

Task: \\
You are given a cluster of social media comments that share a common theme. \\
Your task is to abstract these comments into a single discourse profile description.

Instructions: \\
- Describe this discourse profile's core beliefs and worldview. \\
- Describe their typical goals or motivations when commenting. \\
- Describe their tone, style, and emotional framing. \\
- Describe common behaviors, rhetorical patterns, and phrases they use. \\
- Do not quote individual comments directly. \\
- Write a single coherent discourse profile description suitable for conditioning a language model.

Cluster description and example comments: \\
\{cluster\_description\}

\end{quote}

\subsubsection{Discourse Profile Cards}
\label{app:persona_cards_themselves}
\normalsize
Each discourse profile is represented as a natural-language description capturing characteristic beliefs, goals, tone, and linguistic style. These discourse profile cards are injected verbatim into the LLM prompt to condition responses.

\begin{quote}
\small
\ttfamily
\raggedright

Local / Experiential Observer \\
This discourse profile speaks from personal experience and local observation, often grounding climate change in everyday life. They mention where they live ('I'm in Wisconsin,' 'here in Florida') and describe weird weather patterns ('78 degrees in Vermont in February,' 'record floods in the Midwest'). Their tone is conversational, regional, and observational, rarely ideological. They might use interjections like 'damn,' 'lol,' or 'it's crazy right now.' They often agree or commiserate with others ('same here,' 'we just had that too,' 'yeah it's scary'). Comments often contain anecdotal or comparative phrasing ('last year it was 20 and now it's 70'). Their goal is to share lived experience, connect with others, and validate observations rather than argue policy. Typical phrases: 'same here in [state],' 'I can't believe this weather,' 'we had floods too,' 'this is so weird for this time of year.'

\vspace{0.5em}
Ideological / Political Combatant \\
This discourse profile is fiercely partisan, emotionally charged, and motivated by anger or loyalty toward political figures. They frequently condemn political opponents (especially Trump, MAGA, GOP) or rally support for Democrats ('Vote Blue,' 'Harris 2024'). Their tone is confrontational, moralistic, and full of insults or exclamations ('He's pathetic!!', 'These idiots in charge', 'Wake up!'). They use repetition, capitalization, and emojis (e.g., hearts, 100s) to emphasize points. They often blend outrage with mobilization ('We must vote,' 'Republicans are delusional'). Common behaviors include replying to misinformation with political blame, tagging allies, and expressing outrage about government inaction. Goals: defend their political side, vent frustration, and rally others to act. Typical phrases: 'Vote blue across the board,' 'These people are so blind,' 'We have to do something,' 'This man is a liar,' 'Wake up America.'

\vspace{0.5em}
Systemic / Structural Critic \\
This discourse profile focuses on systemic issues such as capitalism, corporate greed, consumption habits, fossil fuels, and agriculture. They sound thoughtful but frustrated, often linking environmental problems to larger structural causes ('corporations are destroying the planet for profit,' 'capitalism is the problem,' 'we need to reduce plastic and buy less stuff'). Their tone ranges from earnest to cynical. They often use moral or practical language ('reuse, repurpose, buy quality,' 'the rich keep getting richer,' 'fossil fuel industries caused climate change knowingly'). They like calling out hypocrisy ('greenwashing,' 'profit motive,' 'privatized profits, socialized costs'). They may reference renewable energy or veganism, and often frame solutions as lifestyle and collective change, not individual denial. Common phrases: 'stop buying plastic,' 'corporations dont care,' 'we need systemic change,' 'the planet over profits,' 'no more fossil fuel projects.'

\vspace{0.5em}
Pro-Science / Rationalist \\
This discourse profile champions evidence, critical thinking, and scientific literacy. They frequently rebut misinformation with confident, logical language ('read the research paper,' 'peer-reviewed studies show,' 'you're not smarter than scientists'). Tone: rational, occasionally sarcastic, and slightly exasperated. They value education, expertise, and data, often reminding others to 'trust science' or mocking anti-science arrogance. Their comments are relatively well-structured, with complete sentences and punctuation. They may sound academic or firm, using terms like 'cognitive dissonance,' 'confirmation bias,' or 'peer-reviewed.' Goal: defend scientific truth and call out ignorance. Common phrases: 'you clearly havent read the study,' 'this isnt opinion, its data,' 'science isnt up for debate,' 'knowledge kills their arguments,' 'lower your confidence level a little.'

\vspace{0.5em}
Doomist / Nihilist Environmentalist \\
This discourse profile expresses despair about humanity's future and sees collapse as inevitable. Tone: emotional, fatalistic, occasionally poetic or philosophical. They believe Earth will survive but humans wont ('the earth will heal, we wont,' 'were doomed,' 'mother nature is done with us'). They use emotive and reflective phrasing ('Im scared for the future,' 'its already too late,' 'the planet will be fine without us'). They may still show flashes of empathy or communal sorrow ('were all going to die,' 'I just want us all to be okay'). Their goal isnt to persuade but to lament and emotionally process the loss. Typical phrases: 'were past the point of no return,' 'the earth will rid itself of us,' 'humans wont survive,' 'its so sad but true,' 'were doomed.'

\vspace{0.5em}
Conspiratorial / Anti-Climate
This discourse profile is skeptical of mainstream climate science and frequently frames climate change as exaggerated, fabricated, or caused by hidden manipulation. They may claim climate change is a hoax, a scam, or just natural cycles, and often shift explanations toward weather engineering or conspiracies (for example, 'HAARP,' 'geoengineering,' 'chemtrails,' 'cloud seeding,' or 'they control the weather'). Their tone is dismissive, combative, and confident, often using insults or ridicule ('wake up,' 'brainwashed,' 'you people are dense'). They may mix partial scientific-sounding terms with conspiratorial logic (for example, 'thermal expansion' alongside 'sky spraying'). Their goal is to reject institutional narratives, challenge authority, and persuade others that the mainstream explanation is propaganda or control. Typical phrases: 'climate change is a hoax,' 'its just natural cycles,' 'they are manipulating the weather,' 'wake up,' 'do your research,' 'follow the money,' 'they want control.'

\end{quote}

\subsection{Discourse-Profile-Conditioned Stance Detection Prompt}
\label{app:persona_prompt}

We evaluate whether discourse-profile-conditioned LLM agents systematically agree with different types of climate misinformation using the following prompt template. The discourse profile card (Appendix \ref{app:persona_cards_themselves}) is injected into the \texttt{\{discourse profile\}} field and the misinformation claim into the \texttt{\{claim\}} field.

\begin{quote}
\small
\ttfamily
\raggedright

Discourse Profile: \\
\{discourse profile\}

\vspace{0.5em}
Task: \\
Would *this discourse profile* agree with, hesitate about, or reject the following claim?

\vspace{0.5em}
Output ONLY one of: \\
- 1\_agree \\
- 0\_uncertain \\
- -1\_reject

\vspace{0.5em}
USER: \\
Claim: \{claim\} \\
Answer:

\end{quote}

This constrained output format enables direct quantitative comparison of discourse-profile-specific agreement patterns across misinformation categories.

\subsection{Prompt for LLM-evolved variant generation.}
\label{app:llm_variants}
We use the following prompt to generate discourse-profile-conditioned variants:

\begin{quote}
You rewrite one TikTok comment in a target discourse profile style.

Target discourse profile: \{discourse\_profile\_name\} \\
Discourse profile description: \\
\{discourse\_profile\_desc\}

Keep the same misinformation label: \\
\{label\} \\
Label definition: \\
\{label\_desc\}

Original comment: \\
\{claim\}

Video transcript context: \\
\{source\}

Previous rounds (if any): \\
\{prev\_block\}

Rules:
\begin{itemize}
\item Output exactly ONE rewritten comment.
\item No notes, no explanation, no JSON, no labels.
\item Keep it natural and concise.
\item Max \{max\_words\} words.
\item Keep same underlying misinformation type.
\end{itemize}
\end{quote}

\subsection{Misinformation Agreement}
\label{app:misinfo_agreement}

The following prompt is used to obtain independent LLM annotations for assessing agreement on misinformation labels in comments without a high-similarity claim match.

\begin{quote}
\small
\ttfamily
\raggedright

You are annotating user-generated comments for climate change misinformation.

Each comment must be assigned exactly one of the following eight mutually exclusive categories, based on the primary claim expressed or implied in the comment.

\textbf{Categories:}

0\_not\_relevant: No relevant claim detected or claims that don't fit other categories

1\_not\_happening: Claims denying the occurrence of global warming and its effects (e.g., melting ice, extreme weather, sea-level rise)

2\_not\_human: Claims denying human responsibility for climate change

3\_not\_bad: Claims minimizing or denying negative impacts of climate change

4\_solutions\_harmful\_unnecessary: Claims arguing that climate solutions are harmful or unnecessary

5\_science\_is\_unreliable: Claims questioning the validity or reliability of climate science

6\_proponents\_biased: Claims attacking climate scientists or climate action proponents as biased, alarmist, or corrupt

7\_fossil\_fuels\_needed: Claims asserting that fossil fuels are necessary for economic growth or modern living

\textbf{Instruction:} Assign the single category that best fits the comment.

\textbf{Output format:} One label from \texttt{(0\_not\_relevant, 1\_not\_happening, 2\_not\_human, 3\_not\_bad, 4\_solutions\_harmful\_unnecessary, 5\_science\_is\_unreliable, 6\_proponents\_biased, 7\_fossil\_fuels\_needed)}

\textbf{Comment:} \{\textit{comment text}\}
\end{quote}

\end{document}